\documentclass[letterpaper]{article} 
\usepackage{aaai2027}  
\usepackage[hyphens]{url}  
\usepackage{graphicx} 
\usepackage{natbib}  
\usepackage{caption} 
\usepackage{algorithm}
\usepackage{algorithmic}
\usepackage{amsmath}
\usepackage{amssymb}
\usepackage{booktabs}
\usepackage{multirow}
\usepackage{array}
\usepackage[table]{xcolor}
\usepackage{listings}
\usepackage[skins,listings]{tcolorbox}
\usepackage{tikz}
\usepackage[nosfdefault]{comicneue}
\usetikzlibrary{arrows.meta,positioning,fit,backgrounds,calc,decorations.pathmorphing}

\definecolor{elemC}{HTML}{3B3B3B}
\definecolor{elemO}{HTML}{D11919}
\definecolor{elemH}{HTML}{B9B9B9}
\definecolor{elemN}{HTML}{2E5BE6}
\definecolor{elemZn}{HTML}{7B6CB3}   

\newcommand{\zC}{\textcolor{elemC}{C}}
\newcommand{\zO}{\textcolor{elemO}{O}}
\newcommand{\zH}{\textcolor{elemH!60!black}{H}}
\newcommand{\zN}{\textcolor{elemN}{N}}

\definecolor{okgreen}{HTML}{1B7837}
\definecolor{okbg}{HTML}{E8F4EA}
\definecolor{badred}{HTML}{B2182B}
\definecolor{badbg}{HTML}{FBE9E8}
\definecolor{groupbg}{HTML}{F2F2F2}
\definecolor{oursbg}{HTML}{EAF1FB}
\definecolor{evidbg}{HTML}{F4F1E8}
\definecolor{evidline}{HTML}{B8A87A}
\definecolor{toolblue}{HTML}{2C5AA0}
\definecolor{rulegray}{HTML}{888888}
\definecolor{basblue}{HTML}{4E9AC4}   
\definecolor{costlow}{HTML}{D96B6B}
\definecolor{costmid}{HTML}{C73E3A}
\definecolor{costhigh}{HTML}{8F1D2C}
\definecolor{costlowbg}{HTML}{FBE9E8}
\definecolor{costmidbg}{HTML}{F5C1BD}
\definecolor{costhighbg}{HTML}{E58E86}
\newcommand{\cmk}{\textcolor{okgreen}{\checkmark}}

\newcommand{\up}{\,$\uparrow$}
\newcommand{\dn}{\,$\downarrow$}
\newcommand{\drop}[1]{\textcolor{badred}{$\downarrow$#1}}   
\newcommand{\gain}[1]{\textcolor{okgreen}{$\uparrow$#1}}
\newcommand{\best}{\textcolor{okgreen}{\textbf{--}}}
\newcommand{\baselinebest}[1]{\cellcolor{okgreen!12}#1}
\newsavebox{\sweeptbl}

\newcommand{\appendixanchor}[1]{}

\newcommand{\casecrayonrule}[1]{%
\begin{tikzpicture}[baseline]
\draw[#1!78, line width=1.7pt, line cap=round, decorate,
      decoration={random steps,segment length=4pt,amplitude=0.35pt}]
      (0.5pt,0) -- (0.99\linewidth,0);
\draw[#1!38, line width=0.7pt, line cap=round, decorate,
      decoration={random steps,segment length=3pt,amplitude=0.25pt}]
      (1pt,0.35pt) -- (0.985\linewidth,0.35pt);
\end{tikzpicture}\par}
\newcommand{\casecrayonlabel}[2]{%
\tikz[baseline=(caseLabel.base)]{
  \node[inner xsep=1pt,inner ysep=0.2pt,
        font=\comicneue\scriptsize\bfseries,text=#1] (caseLabel) {#2};
  \begin{scope}[on background layer]
    \draw[#1!20,line width=5.2pt,line cap=round,decorate,
          decoration={random steps,segment length=2.8pt,amplitude=0.35pt}]
          ([xshift=-1pt]caseLabel.west) -- ([xshift=1pt]caseLabel.east);
    \draw[#1!13,line width=2pt,line cap=round,decorate,
          decoration={random steps,segment length=2pt,amplitude=0.25pt}]
          ([xshift=-1pt,yshift=0.5pt]caseLabel.west) --
          ([xshift=1pt,yshift=0.5pt]caseLabel.east);
  \end{scope}}}
\newcommand{\casebadge}[2]{%
\tikz[baseline=(caseBadge.base)]{
  \node[inner xsep=4pt,inner ysep=2.3pt,text=white,
        font=\comicneue\scriptsize\bfseries] (caseBadge) {#2};
  \begin{scope}[on background layer]
    \path[fill=#1!90,decorate,
          decoration={random steps,segment length=2.2pt,amplitude=0.45pt}]
          ([xshift=-2pt,yshift=-1pt]caseBadge.south west) --
          ([xshift=2pt,yshift=-1pt]caseBadge.south east) --
          ([xshift=2pt,yshift=1pt]caseBadge.north east) --
          ([xshift=-2pt,yshift=1pt]caseBadge.north west) -- cycle;
    \draw[#1!55,line width=0.7pt,decorate,
          decoration={random steps,segment length=2pt,amplitude=0.3pt}]
          ([xshift=-2pt,yshift=-1pt]caseBadge.south west) --
          ([xshift=2pt,yshift=-1pt]caseBadge.south east);
  \end{scope}}}

\lstdefinestyle{evidencexml}{
  language=XML,
  basicstyle=\ttfamily\fontsize{6.35}{7.10}\selectfont,
  keywordstyle=\color{toolblue}\bfseries,
  identifierstyle=\color{elemC},
  stringstyle=\color{okgreen!78!black},
  commentstyle=\color{rulegray}\itshape,
  emph={net,possible_error_signal,evidence,candidate_interpretations,uncertainty},
  emphstyle=\color{badred}\bfseries,
  backgroundcolor=\color{gray!3},
  frame=single,
  framerule=0.45pt,
  rulecolor=\color{evidline},
  framexleftmargin=5pt,
  framexrightmargin=5pt,
  framextopmargin=4pt,
  framexbottommargin=4pt,
  aboveskip=0pt,
  belowskip=0pt,
  columns=fullflexible,
  keepspaces=true,
  showstringspaces=false,
  breaklines=true,
  breakatwhitespace=false,
  breakindent=1em,
  tabsize=2,
  escapeinside={(*@}{@*)}
}

\nocopyright

\title{\textsc{MOF-Sleuth}: Tool-Grounded Reward Alignment for Explainable Fine-Grained MOF CIF Auditing}
\author{
    Yu Liu\textsuperscript{\rm 1,\rm 2}\equalcontrib,
    Zhiwei Yang\textsuperscript{\rm 1,\rm 2}\equalcontrib,
    Diandian Guo\textsuperscript{\rm 1,\rm 2},
    Kun Peng\textsuperscript{\rm 1,\rm 2},
    Fangfang Yuan\textsuperscript{\rm 1}\corresponding,
    Cong Cao\textsuperscript{\rm 1},\\
    Chaozhuo Li\textsuperscript{\rm 4},
    Zhiyuan Ma\textsuperscript{\rm 5},
    Yanbing Liu\textsuperscript{\rm 1,\rm 2},
    Guobin Zhao\textsuperscript{\rm 3}\corresponding
}
\affiliations{
    \textsuperscript{\rm 1}Institute of Information Engineering, Chinese Academy of Sciences, Beijing, China\\
    \textsuperscript{\rm 2}School of Cyber Security, University of Chinese Academy of Sciences, Beijing, China\\
    \textsuperscript{\rm 3}National University of Singapore, Singapore\\
    \textsuperscript{\rm 4}Beijing University of Posts and Telecommunications, Beijing, China\\
    \textsuperscript{\rm 5}Huazhong University of Science and Technology, Wuhan, China\\
    liuyu@iie.ac.cn, guobinzhao@nus.edu.sg
}

\begin{document}

\maketitle

\begin{abstract}
Large metal-organic framework (MOF) databases support simulation, screening,
and machine learning through crystallographic information files (CIFs). Subtle
chemical and structural errors in these information-dense inputs can
compromise downstream results and hinder manual inspection. Recent LLM
advances in computational chemistry offer paths beyond predictive screening
toward fine-grained diagnosis with evidence-grounded
explanations. However, two challenges remain:
\textbf{\textit{(i) limited fine-grained attribution:}} MOF-specific validators
and machine-learning models scale detection, but provide fixed checks, scalar
readiness scores, or coarse labels rather than evidence-grounded explanations;
and \textbf{\textit{(ii) unreliable CIF reasoning:}} direct LLM auditing is
costly and unreliable because decisive chemical evidence is implicit across
atom-site records and requires geometric, connectivity, occupancy, and charge
calculations. Both stem from weak coupling between computable chemical evidence
and language-model explanation. We introduce
\textbf{\textsc{MOF-Sleuth}}, a reinforcement-guided CIF auditing agent with
two modules: a deterministic \textbf{Forensic Lab} and a
\textbf{Sleuth} reasoning engine. The Lab derives composition, geometry,
connectivity, occupancy, coordination, and charge evidence, and Sleuth uses
this evidence to produce an evidence-grounded explanation, fine-grained error
types, and a binary decision.
Reward-guided reinforcement learning (RL) turns tool measurements into
\textbf{\textit{chemical explanation-level supervision}}, rewarding not only the final
answer but also \textbf{\textit{cited chemical evidence and
evidence-supported diagnoses}}. We introduce Chemically Grounded Diagnosis (Chem-GD), a
deterministic metric that assesses whether a correct diagnosis is explained by
factual, relevant CIF-derived evidence.
Across four benchmarks,
\textbf{\textsc{MOF-Sleuth}} establishes state-of-the-art performance among
evaluated LLM-based approaches and MOF-specific machine-learning methods,
demonstrating gains in detection, attribution, and grounded explanation quality.
\end{abstract}

\begin{figure}[!t]
\centering
\includegraphics[width=0.95\linewidth]{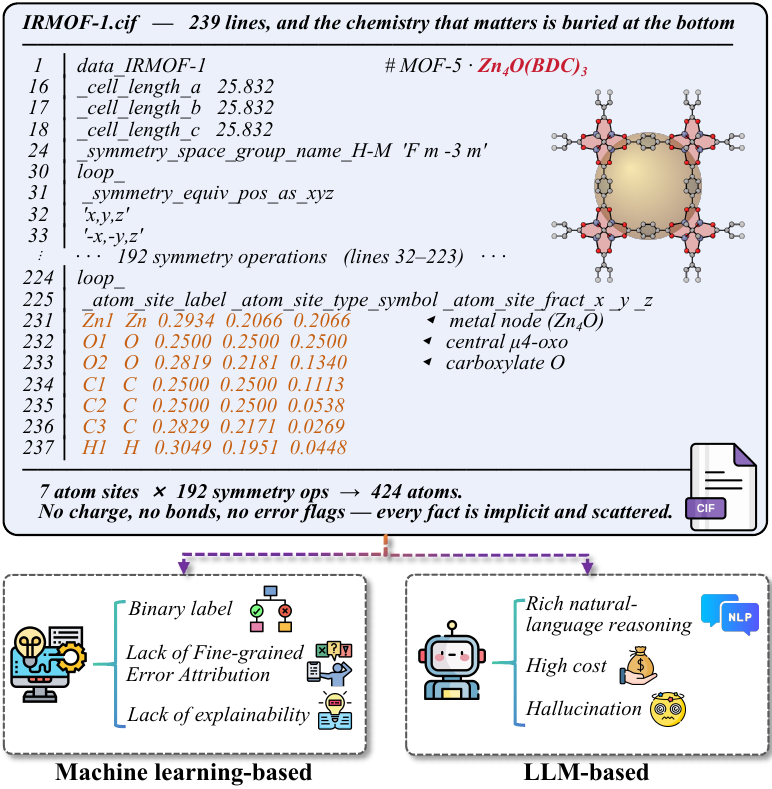}
\caption{MOF structure and CIF serialization. Existing machine-learning
methods offer limited fine-grained error attribution, whereas LLMs struggle to
reason over long, information-dense CIFs whose chemical validity is implicit across
unit-cell fields and atom-site rows.}
\label{fig:mof-cif-representation}
\end{figure}

\section{Introduction}

Metal-organic frameworks (MOFs) are modular porous crystals whose large design
space has driven structure databases and high-throughput simulation and
machine-learning (ML) methods~\cite{furukawa2013mofchemistry,zhao2025coremofdb};
these workflows use crystallographic information files (CIFs) as primary
structural inputs~\cite{hall1991cif}. Yet syntactically readable CIFs are not
necessarily computation-ready: missing atoms, incorrect protonation, charge
imbalance, disorder, abnormal occupancy, or implausible coordination can
silently propagate into downstream results~\cite{gibaldi2025setc}. As modern
MOF collections grow, improving both the efficiency and quality of
CIF error detection becomes critical for reliable screening.

Automated methods have begun to address CIF reliability at scale, but
fine-grained attribution remains limited. MOFChecker~\cite{jin2025mofchecker}
uses rule-based geometric and charge checks with scripted corrections.
MOFClassifier~\cite{zhao2025mofclassifier} predicts computation readiness,
whereas SETC~\cite{gibaldi2025setc} predicts coarse error families. Recently,
LitMOF~\cite{kim2025litmof} introduced an LLM-driven multi-agent workflow for
multi-source MOF curation and structural repair. However, these systems improve
detection, readiness scoring, or repair, but they do not provide a standardized
closed-evidence setting for classifying fine-grained failure modes with
evidence-grounded per-CIF explanations.

LLMs offer a path from validity screening to explainable
diagnosis~\cite{boiko2023coscientist,bran2024chemcrow}. But raw CIFs are
long, table-heavy records whose decisive chemical evidence is implicit across
atom-site fields and derived relations. Validity also depends on objective
geometric, connectivity, occupancy, and charge/protonation
calculations~\cite{gibaldi2025setc}. Without explicit computational grounding,
fluent explanations may therefore rest on
hallucinated or irrelevant evidence~\cite{gao2023alce}. \textit{How can MOF
CIF auditing achieve fine-grained error classification and attribution while
producing evidence-grounded chemical explanations?}

To address this gap, we propose \textsc{MOF-Sleuth}, a reinforcement-guided CIF
auditing agent whose internal workflow separates deterministic evidence
construction from fine-grained diagnosis and explanation. The agent contains
two coordinated modules. A \textbf{Forensic Lab} tool library converts each CIF
into a compact report of objective facts, hard flags,
diagnostic signals, context, and citation aliases; a \textbf{Sleuth} reasoning
policy interprets the report and produces an initial evidence-grounded
explanation, fine-grained error types, and a binary decision. A
verdict-preserving inference stage further refines predicted-error attribution
without altering the initial decision. Unlike
supervised adaptation, which imitates fixed target traces, our setting provides
deterministic verifiers for the final audit. We therefore use reward-guided RL to
\textbf{\textit{optimize why a chemical audit is justified}}: audits are rewarded not
only for correctness and type consistency, but also for whether
\textbf{\textit{LLM explanations cite verified chemical evidence and support
the predicted diagnoses}}. We also define Chemically Grounded Diagnosis (Chem-GD), a
deterministic, class-balanced metric that assesses whether a correct diagnosis
is accompanied by a faithful explanation linking relevant CIF-derived evidence
to its predicted error types. Our contributions are fourfold:

\begin{itemize}
\setlength{\itemsep}{2pt}\setlength{\parskip}{0pt}\setlength{\topsep}{2pt}
\item We present a first systematic formulation of MOF CIF auditing beyond
binary screening, defining it as 15-type fine-grained attribution with
evidence-grounded explanations with RL-guided LLM agent.
\item We propose \textsc{MOF-Sleuth}, a tool-grounded CIF auditing agent whose
Forensic Lab builds deterministic chemical evidence and whose Sleuth policy
outputs decisions, attributions, and evidence-citing explanations.
\item We design evidence-facing rewards that convert deterministic tool outputs
into \textbf{\textit{chemical explanation-level supervision}}, beyond
final-answer correctness.
\item Across four benchmarks and blinded expert validation, \textsc{MOF-Sleuth}
outperforms evaluated LLM-based and MOF-specific baselines; ablations show
complementary gains from tool evidence and reward-guided alignment.
\end{itemize}

\begin{figure*}[t]
\centering
{\includegraphics[width=\textwidth]{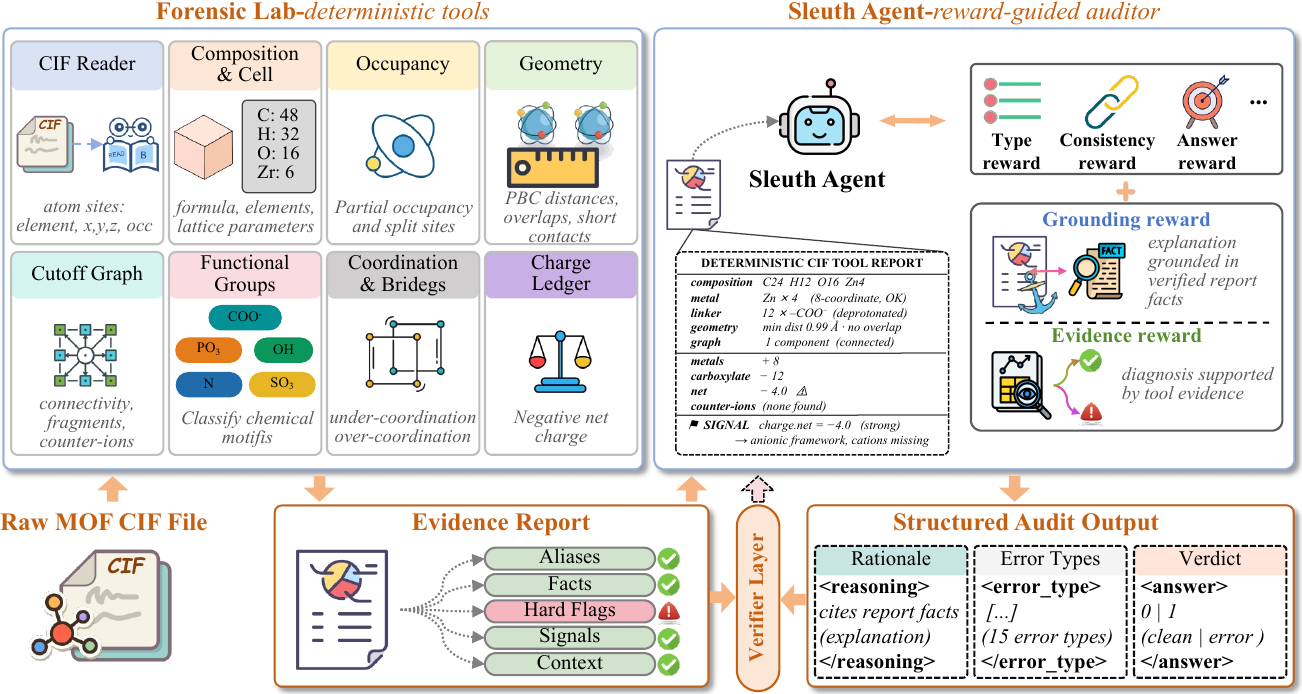}};
\caption{\textbf{Overview of the \textsc{MOF-Sleuth} agent.} Forensic Lab constructs a
structured evidence report from a CIF; Sleuth returns an evidence-grounded
explanation, fine-grained error types, and a binary verdict. Conditional
refinement improves predicted-error attribution while preserving the verdict.}
\label{fig:overview}
\end{figure*}

\section{Related Work}

\noindent\textbf{AI for MOFs and CIF reliability.}
Large-scale MOF discovery relies on computation-ready resources such as
CoRE-MOF~\cite{chung2014coremof,chung2019coremof2019},
ToBaCCo~\cite{colon2017tobacco}, and QMOF~\cite{rosen2021qmof}, but syntactic
readability cannot ensure chemical reliability. Validators and
chemistry-aware models address parts of it: MOFChecker applies
geometric/charge checks~\cite{jin2025mofchecker}, MOSAEC targets metal
oxidation-state inconsistencies~\cite{white2025mosaec}, MOFClassifier predicts
a scalar readiness score~\cite{zhao2025mofclassifier}, SETC classifies proton,
charge, and disorder families~\cite{gibaldi2025setc}, and LitMOF uses
multi-source LLM agents for MOF curation and repair~\cite{kim2025litmof}.
However, these systems emphasize fixed checks, scalar scores, coarse
families, or external-reference repair rather than a common closed-evidence
benchmark with standard verdicts, fine-grained attributions, and
evidence-grounded explanations.

\noindent\textbf{LLMs and agents for scientific reasoning.}
Chain-of-thought~\cite{wei2022chain}, self-consistency~\cite{wang2023selfconsistency},
ReAct~\cite{yao2023react}, Reflexion~\cite{shinn2023reflexion},
AutoGen~\cite{wu2024autogen}, MetaGPT~\cite{hong2024metagpt}, and
DSPy~\cite{khattab2024dspy} improve reasoning, tool use, and workflow
composition. Chemistry agents and models extend this direction to search,
planning, prediction, dialogue, molecular tasks, and text mining~\cite{bran2024chemcrow,
boiko2023coscientist,jablonka2024predictivechemistry,zhao2024chemdfm,
yu2024llasmol,zhang2024chemicaltextmining,kim2025litmof}. However,
closed-evidence CIF auditing requires explanations grounded in distributed
structural evidence from long tables, where fluent generations may still make
unsupported claims~\cite{gao2023alce,min2023factscore}.

\noindent\textbf{Reinforcement learning for structured reasoning.}
Reinforcement learning adapts language models beyond next-token likelihood
through objectives such as PPO~\cite{schulman2017ppo} and GRPO for mathematical
reasoning~\cite{shao2024deepseekmath}. DeepSeek-R1 shows that large-scale RL
can elicit reflection and verification~\cite{guo2025deepseekr1}, and DAPO
studies stable long-CoT RL~\cite{yu2025dapo}. These settings emphasize
general reasoning and final-answer performance, not chemical audits requiring
error attribution, schema consistency, factual grounding, and diagnostic
support.

\section{Methodology}

\subsection{Problem Definition}
\label{sec:problem-definition}

MOF CIF auditing is a closed-evidence structural-diagnosis task. Its label
space $\mathcal{Y}$ comprises 14 named types plus \emph{other} for supported
structural errors outside them. For an input CIF $x$ in the CIF space
$\mathcal{X}$, an auditor predicts an
error-type set $\hat{\mathcal{Y}}\subseteq\mathcal{Y}$ and an evidence-grounded
explanation using only CIF-derived evidence rather than external databases or
literature. Because public MOF error datasets do not provide unified official
labels for these fine-grained causes, the 15-type space is used as an
actionable attribution vocabulary and validated through expert judgment. Free
solvent or guest molecules alone are not errors, so
$\hat{\mathcal{Y}}=\emptyset$.

\begin{figure}[th]
\centering
\includegraphics[width=\columnwidth]{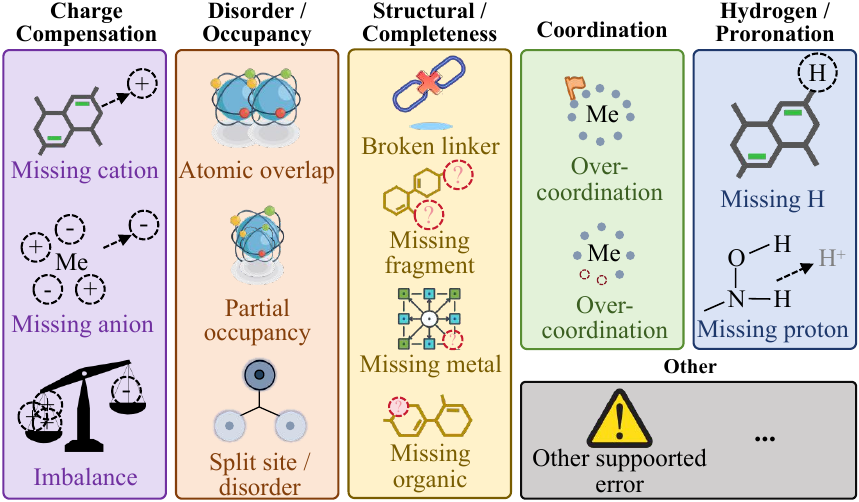}
\caption{The 15 fine-grained CIF error categories.}
\label{fig:audit-label-space}
\end{figure}

\noindent\textbf{Formalization.}
For a deterministic report $r$ derived from $x$, the auditor $f_\theta$, with
parameters $\theta$, returns
\begin{equation}
    f_\theta(r)=(z,\hat{\mathcal{Y}},\hat a).
\end{equation}
Here $z$ is the explanation and $\hat a\in\{0,1\}$ the binary verdict ($1$
erroneous and $0$ clean). With $\mathbb{I}[\cdot]$ denoting an
indicator and $|\cdot|$ set cardinality, audits must satisfy
\begin{equation}
    \hat a=\mathbb{I}\!\left[|\hat{\mathcal{Y}}|>0\right].
\label{eq:type-answer-consistency}
\end{equation}

The 15 fine-grained types are organized under four parent families
$\mathcal{P}=\{\mathrm{charge},\mathrm{hydrogen},\mathrm{disorder},
\mathrm{other}\}$. The fixed multi-label map
$m:\mathcal{Y}\to2^{\mathcal{P}}$ maps each fine type to its parent family or
families, and extends to any set $S\subseteq\mathcal{Y}$ as
$m(S)=\bigcup_{y\in S}m(y)$. Training compares the predicted verdict with the
reference verdict, compares $m(\hat{\mathcal{Y}})$ with available parent-family
annotations when present, and requires $z$ to be grounded in $r$.

\subsection{Framework Overview}

\noindent\textbf{The \textsc{MOF-Sleuth} agent is built from Forensic Lab and
Sleuth.}
As summarized in Figure~\ref{fig:overview}, Forensic Lab converts each CIF into
a structured evidence report, and Sleuth performs semantic reasoning over this
chemical evidence. The output, shown in bottom-right of
Figure~\ref{fig:overview}, is a schema-constrained audit report containing an
evidence-grounded explanation, a subset of the 15-type attribution space, and a
binary verdict; the exact output contract is detailed in
Appendix A.3. Deterministic verification checks
whether the explanation and attribution are supported by the report, and the
same checks supply reward signals, filter verdict-preserving attribution
refinement, and compute Chem-GD.

\noindent\textbf{\textit{Fine-grained attribution}} is (i) emitted as 15-type
labels, (ii) linked to tool diagnostic signals, and (iii) refined or rewarded
only when supported by structural evidence.

\noindent\textbf{\textit{Explanation}} is (i) emitted as evidence-grounded text,
(ii) taught by reward-guided training to cite and use chemical evidence, and
(iii) evaluated by Chem-GD together with the diagnosis.

\subsection{Deterministic Evidence with Forensic Lab}

\noindent\textbf{Forensic Lab converts raw CIFs into structured evidence
reports.}
Forensic Lab takes a raw CIF $x$ and outputs a compact evidence report $r$, the
only structural input read by Sleuth. This keeps calculation-heavy checks out of
the language model. Its guiding principle is to
\textbf{\textit{push audit-relevant operations with objective computational
criteria into Forensic Lab and expose them as citable chemical evidence}}.
Forensic Lab runs CIF parsing,
composition/cell accounting, occupancy analysis, periodic geometry,
cutoff-graph connectivity, functional-group recognition, coordination/bridge
analysis, and charge-ledger checks. These tools compute reproducible facts and
diagnostic signals without setting the final verdict.

\noindent\textbf{The report separates facts, hard flags, signals, context, and
aliases.}
Formally, let $\mathcal{T}=\{T_k\}_{k=1}^{K}$ be the $K$ deterministic
chemistry tools, with $K=8$ in our implementation. For each CIF $x$, they
produce
\begin{equation}
    r = \mathcal{T}(x) =
    \left(\mathcal{F}, \mathcal{H}, \mathcal{S}, \mathcal{C}, \mathcal{A}\right),
\end{equation}
with
\begin{equation}
\begin{aligned}
\mathcal{F} &= \{(p_i,v_i)\} && \text{objective field--value facts},\\
\mathcal{H} &= \{(h_\ell,\tau_\ell,q_\ell)\} && \text{hard flags},\\
\mathcal{S} &= \{(s_j,\tau_j,w_j)\} && \text{diagnostic signals},\\
\mathcal{C} &= \{\cdots\} && \text{interpretation context},\\
\mathcal{A} &= \{a_m\mapsto(p_m,v_m)\} && \text{citation aliases}.
\end{aligned}
\end{equation}
Here $p_i$ is a report field, $v_i$ its value, $h_\ell$ a deterministic flag,
$\tau_\ell\in\mathcal{Y}$ its associated error type, and $q_\ell$ its
supporting cited claim. A hard flag is a deterministic indicator whose
supporting field values directly map to candidate error types. $s_j$ is a
softer structural cue, $\tau_j\in\mathcal{Y}$ its associated error type, and
$w_j$ its evidence strength. $\mathcal{F}$ stores objective measurements such
as atom counts, occupancies, distances, coordination summaries, and
charge-ledger values. $\mathcal{H}$ stores deterministic hard flags, while
$\mathcal{S}$ stores softer error-oriented cues. $\mathcal{C}$ records
non-verdict context such as free solvent and ambiguous motifs, while
$\mathcal{A}$ maps each machine-readable alias $a_m$ to a canonical field
$p_m$ and value $v_m$. The automatically verifiable fact base is
\begin{equation}
    \mathcal{B}(r)=\mathcal{F}\cup
    \{(a_m,v_m):a_m\mapsto(p_m,v_m)\in\mathcal{A}\}.
\end{equation}

\noindent\textbf{The report grounds claims in evidence.}
Thus, cited field--value claims can be grounded directly in the report.
The same report also defines the tool-supported error-type set
\begin{equation}
    \Gamma(r)=
    \{\tau_\ell:(h_\ell,\tau_\ell,q_\ell)\in\mathcal{H}\}
    \cup
    \{\tau_j:(s_j,\tau_j,w_j)\in\mathcal{S}\},
\end{equation}
which records which diagnoses have explicit structural support in the report.
The hard-flag accessor used by the inference protocol is
$\operatorname{hard}(r)=\mathcal{H}$.
Given an audit $o=(z,\hat{\mathcal{Y}},\hat a)$, a deterministic Verifier Layer
therefore checks whether Sleuth's claims are anchored to the Forensic Lab
output:
\begin{equation}
\mathcal{V}(o,r)=
\left[
\mathcal{Q}(z)\subseteq\mathcal{B}(r),\;
\hat{\mathcal{Y}}\subseteq\Gamma(r)
\right],
\end{equation}
where $\mathcal{Q}(z)$ extracts cited claims from the explanation. Schema
validity and type--verdict consistency are checked by the same layer.

\noindent\textbf{The report is reused across training, inference, and
evaluation.}
Forensic Lab supplies reproducible evidence and graded diagnostic cues without
setting a global verdict; Sleuth performs semantic reasoning over chemical
evidence. The
same report structure supports reward computation, verdict-preserving
attribution filtering, and Chem-GD evaluation.
Appendix A.4 details the tools and report fields.

\subsection{Reward-Guided Alignment}

\noindent\textbf{Tool-derived evidence becomes reward supervision for
chemical explanations.}
Computational chemistry audits require correct verdicts, taxonomy-aligned
attributions, and explanations that follow from precise structural evidence.
Prompting can impose the output schema, but cannot ensure that an explanation
cites relevant facts or that the claimed error type is chemically supported.
Forensic Lab therefore supplies the report read by Sleuth and enables
deterministic checks on whether the audit uses that report faithfully. We train
Sleuth with group-relative policy optimization (GRPO)~\cite{shao2024deepseekmath}
using a structured utility whose explanation-facing terms
\textbf{\textit{turn tool-derived evidence into chemical explanation-level
supervision}}: $R_{\mathrm{grd}}$ verifies cited chemical field--value claims
against the report, and $R_{\mathrm{evid}}$ checks predicted error types
against tool-derived structural signals. Thus, $R_{\mathrm{grd}}$ tests whether
cited evidence is real, while $R_{\mathrm{evid}}$ tests whether the diagnosis is
structurally supported. They penalize two audit-critical failure modes in chemical explanations that prompting or
supervised imitation can leave unresolved:
\textbf{\textit{fabricated or irrelevant evidence}} and
\textbf{\textit{plausible but unsupported chemical labels}}.

Let $\mathcal{D}=\{(r_i,\mathcal{P}^{\star}_i,a_i)\}_{i=1}^{N}$ be the $N$-example training
set, where $r_i$ is the Forensic Lab report for CIF $i$,
$\mathcal{P}^{\star}_i$ its reference parent-family
set, and $a_i\in\{0,1\}$ its binary label. For each report, let
$o_{i,g}=(z_{i,g},\hat{\mathcal{Y}}_{i,g},\hat a_{i,g})$ denote candidate audit
$g\in\{1,\ldots,G\}$. We organize six
deterministic verifier rewards by their role before aggregation:
\begin{equation}
\begin{aligned}
\mathbf{r}_{i,g}
={}&\Big(
\underbrace{R_{\mathrm{ans}},R_{\mathrm{type}}}_{\text{task}};
\underbrace{R_{\mathrm{fmt}},R_{\mathrm{cons}}}_{\text{schema}};\\[-0.2em]
&\quad
\underbrace{R_{\mathrm{grd}}}_{\text{ground}};
\underbrace{R_{\mathrm{evid}}}_{\text{diagnosis}}
\Big)_{i,g}^{\top},\\
R_{i,g}
={}&U_{\boldsymbol{\lambda}}
\left(o_{i,g};r_i,\mathcal{P}^{\star}_i,a_i\right)
=\boldsymbol{\lambda}^{\top}\mathbf{r}_{i,g}.
\end{aligned}
\label{eq:structured-audit-utility}
\end{equation}
Here $\mathbf{r}_{i,g}\in\mathbb{R}^{6}$ is the reward vector and
$\boldsymbol{\lambda}=(\lambda_{\mathrm{ans}},\lambda_{\mathrm{type}},
\lambda_{\mathrm{fmt}},\lambda_{\mathrm{cons}},\lambda_{\mathrm{grd}},
\lambda_{\mathrm{evid}})^{\top}\in\mathbb{R}_{\geq0}^{6}$ contains the
weights. The task and schema blocks stabilize verdict correctness,
parent-family agreement, parseability, and type--answer consistency; ground
and diagnosis blocks implement the two explanation-facing checks above. Exact
rules, weights, and GRPO objective are given in
Appendix A.2.

\begin{table*}[t]
\centering
\small
\setlength{\tabcolsep}{3.4pt}
\resizebox{\textwidth}{!}{%
\begin{tabular}{lcccccccc}
\toprule
& \multicolumn{1}{c}{\textit{in-distribution}} & \multicolumn{3}{c}{\textit{out-of-distribution}} & \multicolumn{2}{c}{\textit{detection}} & \multicolumn{2}{c}{\textit{diagnosis}} \\
\cmidrule(lr){2-2}\cmidrule(lr){3-5}\cmidrule(lr){6-7}\cmidrule(lr){8-9}
Method & \mbox{CoRE-MOF~2019} & \mbox{CoRE-MOF~2026} & ToBaCCo & QMOF & Avg.\ Acc\up & Avg.\ Rec\up & \mbox{Type-Hit Acc\up} & \mbox{Chem-GD\up} \\
\midrule
\rowcolor{groupbg}
\multicolumn{9}{l}{\textit{Open-weight end-to-end LLMs}} \\
Qwen3-4B~\cite{yang2025qwen3} & 0.536 & 0.482 & 0.499 & 0.248 & 0.441 & 0.844 & 0.438 & 0.055 \\
Qwen3-8B~\cite{yang2025qwen3} & 0.512 & 0.486 & 0.507 & 0.152 & 0.414 & \baselinebest{0.899} & 0.356 & 0.033 \\
Qwen3-30B-A3B~\cite{yang2025qwen3} & 0.522 & 0.534 & 0.473 & 0.324 & 0.463 & 0.814 & 0.281 & 0.080 \\
Gemma-4-31B~\cite{google2026gemma4} & 0.523 & \baselinebest{0.692} & 0.480 & 0.587 & 0.571 & 0.617 & 0.292 & 0.227 \\
\midrule
\rowcolor{groupbg}
\multicolumn{9}{l}{\textit{API end-to-end LLMs}} \\
DeepSeek-v4-Pro~\cite{deepseek2026v4pro} & 0.607 & 0.585 & 0.571 & 0.554 & 0.579 & 0.749 & 0.425 & 0.117 \\
GPT-5.5~\cite{openai2026gpt55} & \baselinebest{0.686} & 0.683 & \baselinebest{0.585} & 0.839 & \baselinebest{0.698} & 0.413 & \baselinebest{0.617} & \baselinebest{0.418} \\
Claude-Sonnet-4.6~\cite{anthropic2026sonnet46} & 0.512 & 0.571 & 0.535 & \baselinebest{\textbf{0.925}} & 0.636 & 0.169 & 0.460 & 0.304 \\
\midrule
\rowcolor{groupbg}
\multicolumn{9}{l}{\textit{Agent-scaffold pipelines}} \\
Self-Consistency~\cite{wang2023selfconsistency} & 0.520 & 0.494 & 0.507 & 0.210 & 0.433 & 0.883 & 0.423 & 0.037 \\
Reflexion~\cite{shinn2023reflexion} & 0.499 & 0.482 & 0.525 & 0.349 & 0.464 & 0.795 & 0.418 & 0.046 \\
Tree-of-Thoughts~\cite{yao2023tot} & 0.535 & 0.495 & 0.529 & 0.389 & 0.487 & 0.708 & 0.452 & 0.078 \\
LATS~\cite{zhou2024lats} & 0.520 & 0.486 & 0.516 & 0.329 & 0.463 & 0.774 & 0.416 & 0.066 \\
AutoGen~\cite{wu2024autogen} & 0.542 & 0.490 & 0.497 & 0.309 & 0.460 & 0.836 & 0.435 & 0.052 \\
CrewAI~\cite{crewai2024} & 0.522 & 0.524 & 0.468 & 0.357 & 0.468 & 0.770 & 0.404 & 0.044 \\
MetaGPT~\cite{hong2024metagpt} & 0.516 & 0.470 & 0.494 & 0.267 & 0.437 & 0.821 & 0.428 & 0.053 \\
LangGraph~\cite{langgraph2024} & 0.509 & 0.466 & 0.507 & 0.332 & 0.454 & 0.797 & 0.418 & 0.049 \\
DSPy~\cite{khattab2024dspy} & 0.518 & 0.487 & 0.478 & 0.456 & 0.485 & 0.638 & 0.441 & 0.086 \\
GPTSwarm~\cite{zhuge2024gptswarm} & 0.497 & 0.479 & 0.515 & 0.368 & 0.465 & 0.735 & 0.439 & 0.060 \\
AFlow~\cite{zhang2025aflow} & 0.532 & 0.502 & 0.457 & 0.150 & 0.410 & 0.858 & 0.411 & 0.045 \\
AgentSquare~\cite{shang2025agentsquare} & 0.525 & 0.507 & 0.452 & 0.221 & 0.426 & 0.740 & 0.439 & 0.089 \\
ADAS~\cite{hu2025adas} & 0.532 & 0.490 & 0.525 & 0.288 & 0.459 & 0.818 & 0.442 & 0.068 \\
\midrule
\rowcolor{groupbg}
\multicolumn{9}{l}{\textit{Ours}} \\
\rowcolor{oursbg}
\textsc{MOF-Sleuth} w/o RL & 0.660 & 0.674 & 0.631 & 0.305 & 0.568 & \textbf{0.973} & 0.588 & 0.229 \\
\rowcolor{oursbg}
\textbf{\textsc{MOF-Sleuth}} & \textbf{0.768} & \textbf{0.788} & \textbf{0.943} & 0.626 & \textbf{0.781} & 0.929 & \textbf{0.712} & \textbf{0.713} \\
\bottomrule
\end{tabular}
}
\caption{Main results on one in-distribution split and three OOD transfer
suites. Type-Hit Acc measures parent-family attribution. \textbf{Bold} marks
the best result per column, light-green cells mark the strongest baseline, and
arrows ($\uparrow$) indicate that higher is better.}
\label{tab:main-results}
\end{table*}

\subsection{Evaluation of Chemical Explanations}

\noindent\textbf{A sound explanation must connect a correct diagnosis to
relevant CIF-derived evidence.}
Existing explanation-quality metrics~\cite{gao2023alce,min2023factscore,
golovneva2023roscoe,prasad2023receval} often rely on entailment models or LLM
judges, whereas CIF auditing permits deterministic verification. For case $i$,
let $r_i=\mathcal{T}(x_i)$ and let
$\mathcal{B}_i=\mathcal{B}(r_i)$ be the canonical fact base used for
verification; for raw-CIF baselines, this fact base is used only by the
evaluator under the same rules. Let $\mathcal{Q}(z_i)$ be the extractable chemical field--value and
atom claims in explanation $z_i$. \textbf{\textit{A true but irrelevant fact
cannot justify a chemical diagnosis}}; write $q\rightsquigarrow_i y$ when
verified claim $q$ is relevant to predicted type $y$. We define citation
fidelity $g_i$ and attributable evidence--diagnosis linkage $\ell_i$ as
\begin{equation}
\begin{aligned}
g_i&=\mathbb{I}\!\left[\mathcal{Q}(z_i)\neq\emptyset\ \wedge
\mathcal{Q}(z_i)\subseteq\mathcal{B}_i\right],\\
\ell_i&=\mathbb{I}\!\left[\forall y\in\hat{\mathcal{Y}}_i,\,
\exists q\in\mathcal{Q}(z_i)\cap\mathcal{B}_i:
q\rightsquigarrow_i y\right].
\end{aligned}
\end{equation}
Thus, each predicted type must be supported by at least one verified and
diagnosis-relevant cited claim. A frozen support predicate
$\Delta_i(\hat{\mathcal{Y}}_i)$ further checks whether the predicted type set is
compatible with deterministic structural evidence, including competing signals.
Type--answer consistency and support are combined as
\begin{equation}
\begin{aligned}
d_i=\mathbb{I}\big[&
(\hat a_i=0\wedge\hat{\mathcal{Y}}_i=\emptyset)\\
&\vee(\hat a_i=1\wedge\hat{\mathcal{Y}}_i\neq\emptyset
\wedge\Delta_i(\hat{\mathcal{Y}}_i)=1)\big].
\end{aligned}
\end{equation}
Let $\mathcal{I}_{c}=\{i:a_i=c\}$ contain cases of class $c\in\{0,1\}$. Chem-GD is
a \textbf{\textit{class-balanced strict conjunction}} of verdict correctness,
citation fidelity, type--answer consistency, diagnostic support, and
attributable linkage:
\begin{equation}
\mathrm{Chem\text{-}GD}=\frac{1}{2}\sum_{c\in\{0,1\}}
\frac{\sum_{i\in\mathcal{I}_c}\mathbb{I}[\hat a_i=a_i]g_i d_i\ell_i}
{|\mathcal{I}_c|}.
\end{equation}
Thus, an erroneous case cannot pass with an empty type set, and an all-clean
copier is capped at $0.5$ regardless of class imbalance. Chem-GD objectively
measures whether correct diagnoses are accompanied by evidence-grounded
explanations, without a model-based judge.
All methods are scored with the same deterministic parser and frozen predicates;
Appendix A.5 gives component scores, normalization rules,
support predicates, and validation details.

\begin{table*}[t]
\centering
\footnotesize
\sbox{\sweeptbl}{%
\setlength{\tabcolsep}{2.8pt}
\renewcommand{\arraystretch}{1.04}
\begin{tabular}{@{}p{3.20cm}p{3.25cm}cc cc@{}}
\toprule
Setting & Interface/adaptation & Acc\up & $\Delta$ & \mbox{Chem-GD\up} & $\Delta$ \\
\midrule
\rowcolor{groupbg}
\multicolumn{6}{@{}l}{\textit{(a) Architectural effectiveness}} \\
Qwen3-4B & raw CIF, no training & 0.499 & \drop{0.132} & 0.014 & \drop{0.145} \\
Tree-of-Thoughts & raw CIF, agent scaffold & 0.529 & \drop{0.102} & 0.041 & \drop{0.118} \\
GPT-5.5 & raw CIF, API model & 0.585 & \drop{0.046} & 0.369 & \gain{0.210} \\
\textsc{MOF-Sleuth}$_{\mathrm{SFT}}$ & teacher explanation & 0.796 & \gain{0.165} & 0.023 & \drop{0.136} \\
\textsc{MOF-Sleuth}$_{\mathrm{GPT\text{-}5.5}}$ & tool report, API model & 0.894 & \gain{0.263} & 0.519 & \gain{0.360} \\
\midrule
\rowcolor{oursbg}
\textsc{MOF-Sleuth}$_{\mathrm{w/o\ RL}}$ & tool report, frozen& 0.631 & \best & 0.159 & \best \\
\rowcolor{oursbg}
\textsc{MOF-Sleuth}$_{\mathrm{single}}$ & full reward, one pass & 0.943 & \gain{0.312} & 0.640 & \gain{0.481} \\
\rowcolor{oursbg}
\textbf{\textsc{MOF-Sleuth}} & conditional attribution & \textbf{0.943} & \gain{0.312} & \textbf{0.866} & \gain{0.707} \\
\midrule
\rowcolor{groupbg}
\multicolumn{6}{@{}l}{\textit{(b) Reward-guided training}} \\
w/o $R_{\mathrm{type}}$ & w/o label-parent reward & 0.888 & \drop{0.055}$^{\ast}$ & 0.584 & \drop{0.056}$^{\ast}$ \\
w/o $R_{\mathrm{grd}}$ & w/o grounding reward & 0.924 & \drop{0.019}$^{\ast}$ & 0.041 & \drop{0.599}$^{\ast}$ \\
w/o $R_{\mathrm{evid}}$ & w/o diagnostic-support & 0.856 & \drop{0.087}$^{\ast}$ & 0.471 & \drop{0.169}$^{\ast}$ \\
\midrule
\rowcolor{oursbg}
\textsc{MOF-Sleuth}$_{\mathrm{single}}$ & full reward, $\lambda_{\mathrm{evid}}{=}0.90$ & \textbf{0.943} & \best & \textbf{0.640} & \best \\
\midrule
\rowcolor{groupbg}
\multicolumn{6}{@{}l}{\textit{(c) MOF-specific comparison}} \\
MOFChecker~2.0 & geometry/charge checks& 0.710 & \drop{0.233} & N/A & -- \\
MOFClassifier & PU-CGCNN readiness & 0.761 & \drop{0.182} & N/A & -- \\
SETC-GAT$_{\mathrm{atomic}}$ & three graph classifier & 0.785 & \drop{0.158} & N/A & -- \\
\bottomrule
\end{tabular}}
\noindent
\begin{minipage}[t]{0.6\textwidth}
\vspace{0pt}
\resizebox{\linewidth}{!}{\usebox{\sweeptbl}}
\end{minipage}\hfill
\begin{minipage}[t]{0.39\textwidth}
\vspace{0pt}
\centering
\includegraphics[width=\linewidth,trim=7 8 7 7,clip]{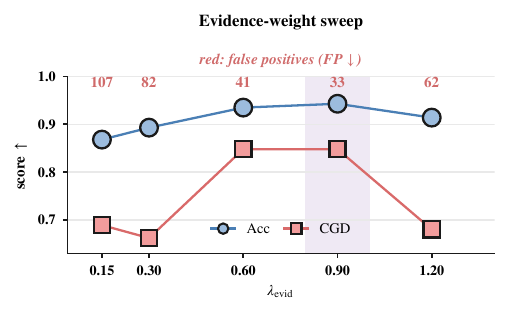}\par
\vspace{2pt}
\includegraphics[width=\linewidth,trim=7 7 7 7,clip]{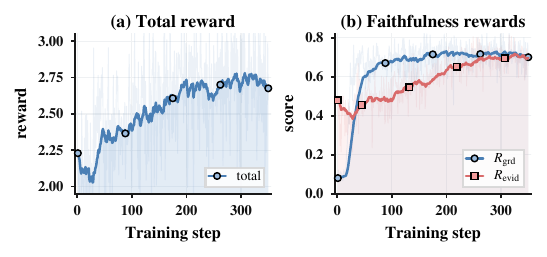}
\end{minipage}\par
\vspace{1pt}
\caption{Ablations and MOF-specific comparisons. The left table
reports architecture, reward, and MOF-specific validator comparisons; the
right-side panels show the evidence-weight sweep (top) and GRPO reward
trajectories (bottom). $\Delta$ denotes change from the corresponding
\textsc{MOF-Sleuth} reference row, and $^{\ast}$ denotes exact McNemar
$p<0.05$.}
\label{tab:ablation}
\end{table*}

\section{Experiments}

\subsection{Experimental Setup}

\noindent\textbf{Datasets.}
We train on human-annotated MOF CIFs and evaluate on CoRE-MOF
2019~\cite{chung2019coremof2019,gibaldi2025setc},
CoRE-MOF 2026~\cite{zhao2025coremofdb}, ToBaCCo~\cite{colon2017tobacco}, and
QMOF~\cite{rosen2021qmof}. These cover in-distribution,
balanced OOD and imbalanced. Split sizes, class
distributions, and sampling procedures are in Appendix C.1.

\noindent\textbf{Baselines.}
We evaluate four groups chosen to cover generic prompting, frontier proprietary
models, reusable agent workflows, and MOF-specific validators.
\textbf{\textit{(i) Open-weight raw-CIF LLMs:}} Qwen3
4B/8B/30B-A3B~\cite{yang2025qwen3} and
Gemma-4-31B~\cite{google2026gemma4} test accessible general models on
serialized CIF text. \textbf{\textit{(ii) Latest flagship API models:}}
DeepSeek-v4-Pro~\cite{deepseek2026v4pro}, GPT-5.5~\cite{openai2026gpt55}, and
Claude-Sonnet-4.6~\cite{anthropic2026sonnet46} test the strongest generic
raw-CIF prompting setting. \textbf{\textit{(iii) General-purpose agent
scaffolds:}} Self-Consistency~\cite{wang2023selfconsistency},
Reflexion~\cite{shinn2023reflexion}, Tree-of-Thoughts~\cite{yao2023tot},
LATS~\cite{zhou2024lats}, AutoGen~\cite{wu2024autogen},
CrewAI~\cite{crewai2024}, MetaGPT~\cite{hong2024metagpt},
LangGraph~\cite{langgraph2024}, DSPy~\cite{khattab2024dspy},
GPTSwarm~\cite{zhuge2024gptswarm}, AFlow~\cite{zhang2025aflow},
AgentSquare~\cite{shang2025agentsquare}, and ADAS~\cite{hu2025adas} test
reusable agent designs on evidence-intensive MOF auditing.
\textbf{\textit{(iv) MOF-specific validators:}} MOFChecker~2.0~\cite{jin2025mofchecker},
MOFClassifier~\cite{zhao2025mofclassifier}, and
SETC-GAT~\cite{gibaldi2025setc} test rule-based and learning-based chemistry
systems.

\noindent\textbf{Metrics.}
All metrics are deterministic. Acc and Avg.\ Rec measure binary
detection, Type-Hit Acc measures four-parent-family attribution, and
class-balanced Chemically Grounded Diagnosis jointly measures diagnostic
correctness and grounded-explanation quality.
Its definition and component analysis are in
Appendices A.5 and C.4.
Targeted ablations include false positives (FP); parseability and type-answer
consistency are auxiliary diagnostics.

\noindent\textbf{Implementation.}
Sleuth is initialized from Qwen3-4B-Instruct~\cite{yang2025qwen3} and trained
with full-parameter GRPO using only the CoRE-MOF 2019 training split. Each
evaluation contributes 1,000 sampled CIFs with manually verified binary
labels. Appendix Figure 6 shows that CoRE-MOF 2026 is
structurally close to the CoRE-MOF reference, whereas QMOF is highly imbalanced
and is therefore property testing with ToBaCCo. Reported results are checked
with repeated complete runs; reward, optimization, decoding, and systems
details are in Appendices A.2 and C.3.

\subsection{Main Results}

\noindent\textbf{Directly reading CIF files is unreliable.}
Table~\ref{tab:main-results} exposes this domain mismatch. Scaling raw-CIF Qwen
models from 4B to 30B leaves average accuracy at $0.414$--$0.463$, while
thirteen generic agent scaffolds remain at $0.410$--$0.487$. All must recover
chemical evidence directly from long, low-level atom-site records, and neither
additional scale nor generic reasoning removes this bottleneck. Even frontier
API models reach only $0.698$ average accuracy under direct CIF reading.
\textsc{MOF-Sleuth} instead leads three of four suites, achieves the strongest
aggregate accuracy ($0.781$), Type-Hit Acc ($0.712$), and Chem-GD ($0.713$),
demonstrating gains in detection, four-family attribution, and grounded
explanation quality. Because QMOF has only 23/1,000 erroneous structures, its
Acc is bias-sensitive (an all-clean classifier reaches $0.977$) and is
interpreted with recall and Chem-GD.

\noindent\textbf{\textsc{MOF-Sleuth} gains come from computable evidence and
reward-guided evidence use.}
With the same frozen 4B auditor, replacing raw CIFs with the Forensic Lab report
raises average accuracy from $0.441$ to $0.568$ and Chem-GD from $0.055$ to
$0.229$, isolating the benefit of deterministic evidence. Reward-guided
alignment teaches the agent to turn this evidence into grounded explanations,
while
verdict-preserving refinement strengthens error attribution without changing
binary decisions. Together, these stages raise average accuracy to $0.781$,
Type-Hit Acc to $0.712$, and Chem-GD to $0.713$. Ablation further
isolates refinement, which raises Chem-GD from $0.640$ to $0.866$ at unchanged
$0.943$ accuracy.

\subsection{Ablations and Comparisons}
Table~\ref{tab:ablation} reports architecture, reward, and domain-validator
  comparisons on ToBaCCo.

\noindent\textbf{Architecture effectiveness.}
Table~\ref{tab:ablation}(a) separates evidence construction from attribution
refinement. With a frozen Qwen3-4B auditor, the Forensic Lab report
beats raw-CIF prompting and Tree-of-Thoughts. The interface also raises
GPT-5.5 Acc/Chem-GD from $0.585/0.369$ to $0.894/0.519$, showing that its benefit
is not backbone-specific. Verdict-preserving refinement increases
Chem-GD from $0.640$ to $0.866$ at unchanged $0.943$ accuracy. These controls
identify evidence construction, rather than model scale or generic reasoning,
as the main bottleneck.

\noindent\textbf{Reward training effectiveness.}
With the evidence interface fixed, SFT reaches $0.796/0.023$ Acc/Chem-GD,
whereas full-reward GRPO reaches $0.943/0.640$. In
Table~\ref{tab:ablation}(b), removing $R_{\mathrm{grd}}$ reduces Chem-GD to
$0.041$, removing $R_{\mathrm{evid}}$ lowers Acc/Chem-GD to $0.856/0.471$, and
removing $R_{\mathrm{type}}$ lowers them to $0.888/0.584$. The
$\lambda_{\mathrm{evid}}$ sweep shows that diagnostic-support weighting controls
weak-signal false positives, and training trajectories show rising total,
grounding, and diagnostic-support rewards. Listed paired differences are significant
($p<0.05$). Thus, $R_{\mathrm{grd}}$ aligns evidence citation, while
$R_{\mathrm{evid}}$ aligns chemically supported labeling.

\noindent\textbf{Comparison with MOF-specific validators.}
Under failure-aware scoring, MOFChecker, MOFClassifier, and SETC-GAT
reach $0.710$--$0.785$ accuracy, compared with $0.943$ for
\textsc{MOF-Sleuth}. These validators do not generate explanations, so Chem-GD is
not applicable to them. Compared with standalone MOF-specific ML validators,
the gain comes from using Forensic Lab as a structured evidence interface and
training Sleuth to integrate deterministic signals into fine-grained,
explainable attributions.

\begin{figure}[t]
\centering
\includegraphics[width=\columnwidth]{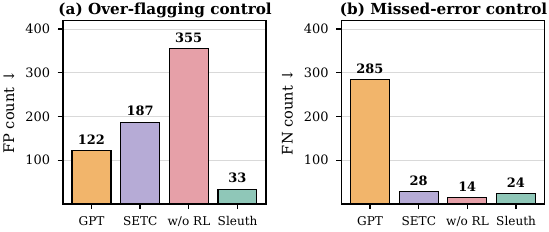}
\caption{Stress-mode comparison, including the strongest MOF-specific ML
baseline. FP measures over-flagging and FN measures missed errors.}
\label{fig:stress-modes}
\end{figure}

\begin{table}[t!]
\centering
\scriptsize
\setlength{\tabcolsep}{3.0pt}
\renewcommand{\arraystretch}{1.08}
\resizebox{\columnwidth}{!}{%
\begin{tabular}{lcccc}
\toprule
Method & Avg.\ Acc\up & Tokens (M)\dn &
s/ex.$\downarrow$ & Cost$\downarrow$ \\
\midrule
DeepSeek-v4-Pro &
\cellcolor{badred!12}0.579 & \cellcolor{badred!18}70.81 &
\cellcolor{badred!15}11.95 &
\cellcolor{costlowbg}\textcolor{costlow}{\$41.89} \\
GPT-5.5 &
\cellcolor{okgreen!12}0.698 & \cellcolor{badred!24}72.83 &
\cellcolor{okgreen!10}4.62 &
\cellcolor{costhighbg}\textcolor{costhigh}{\textbf{\$919.08}} \\
Claude-Sonnet-4.6 &
\cellcolor{okgreen!6}0.636 & \cellcolor{okgreen!12}49.86 &
\cellcolor{badred!20}13.82 &
\cellcolor{costmidbg}\textcolor{costmid}{\$169.43} \\
\midrule
\textbf{\textsc{MOF-Sleuth}} & \cellcolor{okgreen!30}\textbf{0.781} &
\cellcolor{okgreen!30}\textbf{17.50} & \cellcolor{okgreen!24}\textbf{2.04} &
\cellcolor{okgreen!24}\$0.00 \\
\bottomrule
\end{tabular}}
\caption{Efficiency comparison. \textsc{MOF-Sleuth} uses fewer tokens, runs
faster, and avoids API cost.}
\label{tab:api-efficiency}
\end{table}

\subsection{Analysis}

\noindent\textbf{Evidence and alignment address complementary errors.}
Figure~\ref{fig:stress-modes} isolates the precision--recall trade-off on
the same transfer suite. GPT-5.5 keeps false positives moderate but misses 285 errors
on the same 500 erroneous cases. SETC-GAT reduces missed errors to 28 but still
flags 187 clean structures. The frozen tool-report auditor has only 14 FN but
over-flags 355 clean structures, showing that tool evidence improves
sensitivity but needs calibration. Reward-guided alignment keeps missed errors
low (24 FN) while reducing false positives to 33.
\textbf{\textit{RL turns tool sensitivity into calibrated evidence use.}}
This performance gain mainly reflects the evidence-support reward:
$R_{\mathrm{evid}}$ discourages weak-signal over-flagging while preserving the
sensitivity provided by the tool report. The grounding reward is evaluated
separately through Chem-GD and expert-judge validation.

\noindent\textbf{Compact evidence improves the accuracy--cost trade-off for
scalable auditing.}
Using compact reports and a local 4B auditor, \textsc{MOF-Sleuth} reaches
$0.781$ average accuracy with 17.50M tokens and 2.04 s/example
(Table~\ref{tab:api-efficiency}). API models reach $0.579$--$0.698$ using
49.86--72.83M tokens and 4.62--13.82 s/example, costing \$42--\$919 for 4,000
audits under recorded OpenRouter prices~\cite{openrouter2026pricing,
openrouter2026reasoning}. Our token total includes all refinement calls; the
latency conservatively includes full-cohort refinement time although only
53.2\% of cases trigger it.

\subsection{Expert-Judge Validation}

\noindent\textbf{Experts confirm reliable verdicts and useful causes.}
We use blinded expert review to test whether the audit can support CIF
curation. As shown in Table~\ref{tab:expert-judge-main},
\textsc{MOF-Sleuth} reaches 0.930 verdict agreement ($\kappa=0.860$;
Cohen's kappa~\cite{cohen1960agreement}), and its
15-type attributions reach 0.700 Type-Hit Acc on erroneous cases
(Jaccard 0.294~\cite{jaccard1901etude}). This directly supports the task
formulation: the model does not only flag erroneous CIFs, but also gives
chemically meaningful
\textbf{\textit{fine-grained attribution}}.

\begin{table}[t]
\centering
\scriptsize
\setlength{\tabcolsep}{5.0pt}
\renewcommand{\arraystretch}{1.10}
\resizebox{\columnwidth}{!}{%
\begin{tabular}{@{}ll@{}}
\toprule
\textbf{Audit claim} & \textbf{Blinded expert evidence} \\
\midrule
Reliable verdict &
\textcolor{okgreen}{\textbf{0.930}} agreement, $\kappa{=}\textcolor{okgreen}{\textbf{0.860}}$ \\
\textbf{Fine-grained attribution} &
\textcolor{okgreen}{\textbf{0.700}} Type-Hit Acc, Jaccard $=\textcolor{okgreen}{\textbf{0.294}}$ \\
\textbf{Attribution refinement} &
Type-Hit Acc \textcolor{okgreen}{\textbf{+0.240}}, Jaccard \textcolor{okgreen}{\textbf{+0.091}} \\
Chem-GD validity &
Chem-GD precision $=\textcolor{okgreen}{\textbf{0.938}}$, agreement $=\textcolor{okgreen}{\textbf{0.720}}$ \\
\bottomrule
\end{tabular}}
\caption{Expert validation. Blinded review supports verdict reliability,
fine-grained attribution, refinement, and Chem-GD; details are in Appendix D.}
\label{tab:expert-judge-main}
\end{table}

\noindent\textbf{Experts also support the explanation-centered design.}
Verdict-preserving refinement improves Type-Hit Acc by +0.240 and Jaccard by
+0.091 without changing the binary decision, showing that the second pass
improves why the CIF is wrong rather than merely shifting the verdict. Unlike
generic citation or rationale metrics~\cite{gao2023alce,min2023factscore,
golovneva2023roscoe,prasad2023receval}, Chem-GD checks CIF-derived evidence and
diagnosis together; it tracks expert-supported diagnoses (precision 0.938,
agreement 0.720), supporting its use as a scalable, deterministic measure of
evidence-grounded chemical explanations.

\section{Conclusion}

We present the first systematic LLM-based framework for MOF CIF auditing beyond
binary validity screening. \textsc{MOF-Sleuth} combines a deterministic
Forensic Lab, which converts raw CIFs into citable chemical evidence, with a
Sleuth reasoning agent that produces binary verdicts, fine-grained error
attributions, and evidence-grounded explanations. Expert validation supports
the usefulness of its fine-grained attribution and explanation quality, while
large-scale experiments show that reward-guided alignment teaches the agent to
use tool-derived evidence more reliably. Ablations further verify the
architecture and reward design, and \textsc{MOF-Sleuth} leads MOF-specific ML,
API, and agent baselines.

\bibliography{references}

\begin{thebibliography}{50}
\providecommand{\natexlab}[1]{#1}

\bibitem[{{Anthropic}(2026)}]{anthropic2026sonnet46}
{Anthropic}. 2026.
\newblock Introducing {Claude Sonnet 4.6}.
\newblock \url{https://www.anthropic.com/news/claude-sonnet-4-6}.
\newblock Accessed: 2026-06-23.

\bibitem[{Boiko et~al.(2023)Boiko, MacKnight, Kline, and Gomes}]{boiko2023coscientist}
Boiko, D.~A.; MacKnight, R.; Kline, B.; and Gomes, G. 2023.
\newblock Autonomous Chemical Research with Large Language Models.
\newblock \emph{Nature}, 624(7992): 570--578.

\bibitem[{Bran et~al.(2024)Bran, Cox, Schilter, Baldassari, White, and Schwaller}]{bran2024chemcrow}
Bran, A.~M.; Cox, S.; Schilter, O.; Baldassari, C.; White, A.~D.; and Schwaller, P. 2024.
\newblock Augmenting Large Language Models with Chemistry Tools.
\newblock \emph{Nature Machine Intelligence}, 6(5): 525--535.

\bibitem[{Chung et~al.(2014)Chung, Camp, Haranczyk, Sikora, Bury, Krungleviciute, Yildirim, Farha, Sholl, and Snurr}]{chung2014coremof}
Chung, Y.~G.; Camp, J.; Haranczyk, M.; Sikora, B.~J.; Bury, W.; Krungleviciute, V.; Yildirim, T.; Farha, O.~K.; Sholl, D.~S.; and Snurr, R.~Q. 2014.
\newblock Computation-Ready, Experimental Metal--Organic Frameworks: A Tool To Enable High-Throughput Screening of Nanoporous Crystals.
\newblock \emph{Chemistry of Materials}, 26(21): 6185--6192.

\bibitem[{Chung et~al.(2019)Chung, Haldoupis, Bucior, Haranczyk, Lee, Zhang, Vogiatzis, Milisavljevic, Ling, Camp, Slater, Siepmann, Sholl, and Snurr}]{chung2019coremof2019}
Chung, Y.~G.; Haldoupis, E.; Bucior, B.~J.; Haranczyk, M.; Lee, S.; Zhang, H.; Vogiatzis, K.~D.; Milisavljevic, M.; Ling, S.; Camp, J.~S.; Slater, B.; Siepmann, J.~I.; Sholl, D.~S.; and Snurr, R.~Q. 2019.
\newblock Advances, Updates, and Analytics for the Computation-Ready, Experimental Metal--Organic Framework Database: {CoRE MOF} 2019.
\newblock \emph{Journal of Chemical \& Engineering Data}, 64(12): 5985--5998.

\bibitem[{Cohen(1960)}]{cohen1960agreement}
Cohen, J. 1960.
\newblock A Coefficient of Agreement for Nominal Scales.
\newblock \emph{Educational and Psychological Measurement}, 20(1): 37--46.

\bibitem[{Col{\'o}n, G{\'o}mez-Gualdr{\'o}n, and Snurr(2017)}]{colon2017tobacco}
Col{\'o}n, Y.~J.; G{\'o}mez-Gualdr{\'o}n, D.~A.; and Snurr, R.~Q. 2017.
\newblock Topologically Guided, Automated Construction of Metal--Organic Frameworks and Their Evaluation for Energy-Related Applications.
\newblock \emph{Crystal Growth \& Design}, 17(11): 5801--5810.

\bibitem[{{CrewAI Inc.}(2024)}]{crewai2024}
{CrewAI Inc.} 2024.
\newblock {CrewAI}: Framework for Orchestrating Role-Playing Autonomous {AI} Agents.
\newblock \url{https://github.com/crewAIInc/crewAI}.

\bibitem[{{DeepSeek-AI}(2026)}]{deepseek2026v4pro}
{DeepSeek-AI}. 2026.
\newblock {DeepSeek V4} Preview Release.
\newblock \url{https://api-docs.deepseek.com/news/news260424/}.
\newblock Accessed: 2026-06-15.

\bibitem[{Furukawa et~al.(2013)Furukawa, Cordova, O'Keeffe, and Yaghi}]{furukawa2013mofchemistry}
Furukawa, H.; Cordova, K.~E.; O'Keeffe, M.; and Yaghi, O.~M. 2013.
\newblock The Chemistry and Applications of Metal-Organic Frameworks.
\newblock \emph{Science}, 341(6149): 1230444.

\bibitem[{Gao et~al.(2023)Gao, Yen, Yu, and Chen}]{gao2023alce}
Gao, T.; Yen, H.; Yu, J.; and Chen, D. 2023.
\newblock Enabling Large Language Models to Generate Text with Citations.
\newblock In \emph{Proceedings of the 2023 Conference on Empirical Methods in Natural Language Processing}, 6465--6488. Singapore: Association for Computational Linguistics.

\bibitem[{Gibaldi et~al.(2025)Gibaldi, Luo, White, Mayo, Pereira, and Woo}]{gibaldi2025setc}
Gibaldi, M.; Luo, J.; White, A.~J.; Mayo, R.~A.; Pereira, C.; and Woo, T.~K. 2025.
\newblock Generalizable classification of crystal structure error types using graph attention networks.
\newblock \emph{Journal of Materials Chemistry A}, 13: 32255--32270.

\bibitem[{Golovneva et~al.(2023)Golovneva, Chen, Poff, Corredor, Zettlemoyer, Fazel-Zarandi, and Celikyilmaz}]{golovneva2023roscoe}
Golovneva, O.; Chen, M.~P.; Poff, S.; Corredor, M.; Zettlemoyer, L.; Fazel-Zarandi, M.; and Celikyilmaz, A. 2023.
\newblock {ROSCOE}: A Suite of Metrics for Scoring Step-by-Step Reasoning.
\newblock In \emph{International Conference on Learning Representations}.

\bibitem[{{Google}(2026)}]{google2026gemma4}
{Google}. 2026.
\newblock {Gemma 4} Model Overview.
\newblock \url{https://ai.google.dev/gemma/docs/core/model_card_4}.
\newblock Accessed: 2026-06-15.

\bibitem[{Guo et~al.(2025)Guo, Yang, Zhang, Song, Wang, Zhu, Xu, Zhang, Ma, Bi, Zhang, Yu, Wu, Wu, Gou, Shao, Li, Gao, Liu et~al.}]{guo2025deepseekr1}
Guo, D.; Yang, D.; Zhang, H.; Song, J.; Wang, P.; Zhu, Q.; Xu, R.; Zhang, R.; Ma, S.; Bi, X.; Zhang, X.; Yu, X.; Wu, Y.; Wu, Z.~F.; Gou, Z.; Shao, Z.; Li, Z.; Gao, Z.; Liu, A.; et~al. 2025.
\newblock {DeepSeek-R1} Incentivizes Reasoning in {LLMs} through Reinforcement Learning.
\newblock \emph{Nature}, 645(8081): 633--638.

\bibitem[{Hall, Allen, and Brown(1991)}]{hall1991cif}
Hall, S.~R.; Allen, F.~H.; and Brown, I.~D. 1991.
\newblock The Crystallographic Information File ({CIF}): A New Standard Archive File for Crystallography.
\newblock \emph{Acta Crystallographica Section A}, 47(6): 655--685.

\bibitem[{Hong et~al.(2024)Hong, Zhuge, Chen, Zheng, Cheng, Wang, Zhang, Wang, Yau, Lin, Zhou, Ran, Xiao, Wu, and Schmidhuber}]{hong2024metagpt}
Hong, S.; Zhuge, M.; Chen, J.; Zheng, X.; Cheng, Y.; Wang, J.; Zhang, C.; Wang, Z.; Yau, S.; Lin, Z.; Zhou, L.; Ran, C.; Xiao, L.; Wu, C.; and Schmidhuber, J. 2024.
\newblock {MetaGPT}: Meta Programming for A Multi-Agent Collaborative Framework.
\newblock In \emph{International Conference on Learning Representations}, volume 2024, 23247--23275.

\bibitem[{Hu, Lu, and Clune(2025)}]{hu2025adas}
Hu, S.; Lu, C.; and Clune, J. 2025.
\newblock Automated Design of Agentic Systems.
\newblock In \emph{International Conference on Learning Representations}, volume 2025, 21344--21377.

\bibitem[{Jablonka et~al.(2024)Jablonka, Schwaller, Ortega-Guerrero, and Smit}]{jablonka2024predictivechemistry}
Jablonka, K.~M.; Schwaller, P.; Ortega-Guerrero, A.; and Smit, B. 2024.
\newblock Leveraging Large Language Models for Predictive Chemistry.
\newblock \emph{Nature Machine Intelligence}, 6(2): 161--169.

\bibitem[{Jaccard(1901)}]{jaccard1901etude}
Jaccard, P. 1901.
\newblock {\'E}tude comparative de la distribution florale dans une portion des Alpes et des Jura.
\newblock \emph{Bull Soc Vaudoise Sci Nat}, 37: 547--579.

\bibitem[{Jin et~al.(2025)Jin, Jablonka, Moubarak, Li, and Smit}]{jin2025mofchecker}
Jin, X.; Jablonka, K.~M.; Moubarak, E.; Li, Y.; and Smit, B. 2025.
\newblock {MOFChecker}: A Package for Validating and Correcting Metal--Organic Framework ({MOF}) Structures.
\newblock \emph{Digital Discovery}, 4(6): 1560--1569.

\bibitem[{Khattab et~al.(2024)Khattab, Singhvi, Maheshwari, Zhang, Santhanam, A, Haq, Sharma, Joshi, Moazam, Miller, Zaharia, and Potts}]{khattab2024dspy}
Khattab, O.; Singhvi, A.; Maheshwari, P.; Zhang, Z.; Santhanam, K.; A, S.~V.; Haq, S.; Sharma, A.; Joshi, T.; Moazam, H.; Miller, H.; Zaharia, M.; and Potts, C. 2024.
\newblock {DSPy}: Compiling Declarative Language Model Calls into Self-Improving Pipelines.
\newblock In \emph{International Conference on Learning Representations}, volume 2024, 54928--54958.

\bibitem[{Kim, Kim, and Kim(2025)}]{kim2025litmof}
Kim, H.; Kim, D.; and Kim, J. 2025.
\newblock {LLM}-Driven Multi-Agent Curation and Expansion of Metal--Organic Frameworks Database.
\newblock arXiv:2512.01693.

\bibitem[{{LangChain}(2024)}]{langgraph2024}
{LangChain}. 2024.
\newblock {LangGraph}: Build Stateful, Multi-Actor Applications with {LLMs}.
\newblock \url{https://github.com/langchain-ai/langgraph}.

\bibitem[{Min et~al.(2023)Min, Krishna, Lyu, Lewis, Yih, Koh, Iyyer, Zettlemoyer, and Hajishirzi}]{min2023factscore}
Min, S.; Krishna, K.; Lyu, X.; Lewis, M.; Yih, W.-t.; Koh, P.~W.; Iyyer, M.; Zettlemoyer, L.; and Hajishirzi, H. 2023.
\newblock {FActScore}: Fine-grained Atomic Evaluation of Factual Precision in Long Form Text Generation.
\newblock In \emph{Proceedings of the 2023 Conference on Empirical Methods in Natural Language Processing}, 12076--12100. Singapore: Association for Computational Linguistics.

\bibitem[{{OpenAI}(2026)}]{openai2026gpt55}
{OpenAI}. 2026.
\newblock {GPT-5.5} API Model.
\newblock \url{https://developers.openai.com/api/docs/models/gpt-5.5}.
\newblock Accessed: 2026-06-19.

\bibitem[{{OpenRouter}(2026{\natexlab{a}})}]{openrouter2026pricing}
{OpenRouter}. 2026{\natexlab{a}}.
\newblock {OpenRouter} Model Catalog and Pricing.
\newblock \url{https://openrouter.ai/pricing}.
\newblock Accessed: 2026-06-23.

\bibitem[{{OpenRouter}(2026{\natexlab{b}})}]{openrouter2026reasoning}
{OpenRouter}. 2026{\natexlab{b}}.
\newblock {OpenRouter} Reasoning Tokens.
\newblock \url{https://openrouter.ai/docs/guides/best-practices/reasoning-tokens}.
\newblock Accessed: 2026-06-23.

\bibitem[{Prasad et~al.(2023)Prasad, Saha, Zhou, and Bansal}]{prasad2023receval}
Prasad, A.; Saha, S.; Zhou, X.; and Bansal, M. 2023.
\newblock {ReCEval}: Evaluating Reasoning Chains via Correctness and Informativeness.
\newblock In \emph{Proceedings of the 2023 Conference on Empirical Methods in Natural Language Processing}, 10066--10086. Singapore: Association for Computational Linguistics.

\bibitem[{Rosen et~al.(2021)Rosen, Iyer, Ray, Yao, Aspuru-Guzik, Gagliardi, Notestein, and Snurr}]{rosen2021qmof}
Rosen, A.~S.; Iyer, S.~M.; Ray, D.; Yao, Z.; Aspuru-Guzik, A.; Gagliardi, L.; Notestein, J.~M.; and Snurr, R.~Q. 2021.
\newblock Machine Learning the Quantum-Chemical Properties of Metal--Organic Frameworks for Accelerated Materials Discovery.
\newblock \emph{Matter}, 4(5): 1578--1597.

\bibitem[{Schulman et~al.(2017)Schulman, Wolski, Dhariwal, Radford, and Klimov}]{schulman2017ppo}
Schulman, J.; Wolski, F.; Dhariwal, P.; Radford, A.; and Klimov, O. 2017.
\newblock Proximal Policy Optimization Algorithms.
\newblock arXiv:1707.06347.

\bibitem[{Shang et~al.(2025)Shang, Li, Zhao, Ma, Liu, Xu, and Li}]{shang2025agentsquare}
Shang, Y.; Li, Y.; Zhao, K.; Ma, L.; Liu, J.; Xu, F.; and Li, Y. 2025.
\newblock {AgentSquare}: Automatic {LLM} Agent Search in Modular Design Space.
\newblock In \emph{International Conference on Learning Representations}, volume 2025, 3841--3865.

\bibitem[{Shao et~al.(2024)Shao, Wang, Zhu, Xu, Song, Bi, Zhang, Zhang, Li, Wu, and Guo}]{shao2024deepseekmath}
Shao, Z.; Wang, P.; Zhu, Q.; Xu, R.; Song, J.; Bi, X.; Zhang, H.; Zhang, M.; Li, Y.~K.; Wu, Y.; and Guo, D. 2024.
\newblock {DeepSeekMath}: Pushing the Limits of Mathematical Reasoning in Open Language Models.
\newblock arXiv:2402.03300.

\bibitem[{Shinn et~al.(2023)Shinn, Cassano, Gopinath, Narasimhan, and Yao}]{shinn2023reflexion}
Shinn, N.; Cassano, F.; Gopinath, A.; Narasimhan, K.; and Yao, S. 2023.
\newblock Reflexion: Language Agents with Verbal Reinforcement Learning.
\newblock In \emph{Advances in Neural Information Processing Systems}, volume~36, 8634--8652. Curran Associates, Inc.

\bibitem[{Wang et~al.(2023)Wang, Wei, Schuurmans, Le, Chi, Narang, Chowdhery, and Zhou}]{wang2023selfconsistency}
Wang, X.; Wei, J.; Schuurmans, D.; Le, Q.~V.; Chi, E.~H.; Narang, S.; Chowdhery, A.; and Zhou, D. 2023.
\newblock Self-Consistency Improves Chain of Thought Reasoning in Language Models.
\newblock In \emph{International Conference on Learning Representations}.

\bibitem[{Wei et~al.(2022)Wei, Wang, Schuurmans, Bosma, Ichter, Xia, Chi, Le, and Zhou}]{wei2022chain}
Wei, J.; Wang, X.; Schuurmans, D.; Bosma, M.; Ichter, B.; Xia, F.; Chi, E.~H.; Le, Q.~V.; and Zhou, D. 2022.
\newblock Chain-of-Thought Prompting Elicits Reasoning in Large Language Models.
\newblock In \emph{Advances in Neural Information Processing Systems}, volume~35, 24824--24837. Curran Associates, Inc.

\bibitem[{White et~al.(2025)White, Gibaldi, Burner, Mayo, and Woo}]{white2025mosaec}
White, A.~J.; Gibaldi, M.; Burner, J.; Mayo, R.~A.; and Woo, T.~K. 2025.
\newblock High Structural Error Rates in ``Computation-Ready'' {MOF} Databases Discovered by Checking Metal Oxidation States.
\newblock \emph{Journal of the American Chemical Society}, 147(21): 17579--17583.

\bibitem[{Wu et~al.(2024)Wu, Bansal, Zhang, Wu, Li, Zhu, Jiang, Zhang, Zhang, Liu, Awadallah, White, Burger, and Wang}]{wu2024autogen}
Wu, Q.; Bansal, G.; Zhang, J.; Wu, Y.; Li, B.; Zhu, E.; Jiang, L.; Zhang, X.; Zhang, S.; Liu, J.; Awadallah, A.~H.; White, R.~W.; Burger, D.; and Wang, C. 2024.
\newblock {AutoGen}: Enabling Next-Gen {LLM} Applications via Multi-Agent Conversations.
\newblock In \emph{First Conference on Language Modeling}.

\bibitem[{Yang et~al.(2025)Yang, Li, Yang, Zhang, Hui, Zheng, Yu, Gao, Huang, Lv et~al.}]{yang2025qwen3}
Yang, A.; Li, A.; Yang, B.; Zhang, B.; Hui, B.; Zheng, B.; Yu, B.; Gao, C.; Huang, C.; Lv, C.; et~al. 2025.
\newblock {Qwen3} Technical Report.
\newblock arXiv:2505.09388.

\bibitem[{Yao et~al.(2023{\natexlab{a}})Yao, Yu, Zhao, Shafran, Griffiths, Cao, and Narasimhan}]{yao2023tot}
Yao, S.; Yu, D.; Zhao, J.; Shafran, I.; Griffiths, T.; Cao, Y.; and Narasimhan, K. 2023{\natexlab{a}}.
\newblock Tree of Thoughts: Deliberate Problem Solving with Large Language Models.
\newblock In \emph{Advances in Neural Information Processing Systems}, volume~36, 11809--11822. Curran Associates, Inc.

\bibitem[{Yao et~al.(2023{\natexlab{b}})Yao, Zhao, Yu, Du, Shafran, Narasimhan, and Cao}]{yao2023react}
Yao, S.; Zhao, J.; Yu, D.; Du, N.; Shafran, I.; Narasimhan, K.~R.; and Cao, Y. 2023{\natexlab{b}}.
\newblock {ReAct}: Synergizing Reasoning and Acting in Language Models.
\newblock In \emph{International Conference on Learning Representations}.

\bibitem[{Yu et~al.(2024)Yu, Baker, Chen, Ning, and Sun}]{yu2024llasmol}
Yu, B.; Baker, F.~N.; Chen, Z.; Ning, X.; and Sun, H. 2024.
\newblock {LlaSMol}: Advancing Large Language Models for Chemistry with a Large-Scale, Comprehensive, High-Quality Instruction Tuning Dataset.
\newblock In \emph{First Conference on Language Modeling}.

\bibitem[{Yu et~al.(2025)Yu, Zhang, Zhu, Yuan, Zuo, Yue, Dai, Fan, Liu, Liu, Liu, Liu, Lin, Lin et~al.}]{yu2025dapo}
Yu, Q.; Zhang, Z.; Zhu, R.; Yuan, Y.; Zuo, X.; Yue, Y.; Dai, W.; Fan, T.; Liu, G.; Liu, J.; Liu, L.; Liu, X.; Lin, H.; Lin, Z.; et~al. 2025.
\newblock {DAPO}: An Open-Source {LLM} Reinforcement Learning System at Scale.
\newblock In \emph{Advances in Neural Information Processing Systems}, volume~38, 113222--113244. Curran Associates, Inc.

\bibitem[{Zhang et~al.(2025)Zhang, Xiang, Yu, Teng, Chen, Chen, Zhuge, Cheng, Hong, Wang, Zheng, Liu, Luo, and Wu}]{zhang2025aflow}
Zhang, J.; Xiang, J.; Yu, Z.; Teng, F.; Chen, X.; Chen, J.; Zhuge, M.; Cheng, X.; Hong, S.; Wang, J.; Zheng, B.; Liu, B.; Luo, Y.; and Wu, C. 2025.
\newblock {AFlow}: Automating Agentic Workflow Generation.
\newblock In \emph{International Conference on Learning Representations}, volume 2025, 34040--34077.

\bibitem[{Zhang et~al.(2024)Zhang, Wang, Kong, Xiong, Ni, Cao, Niu, Chen, Li, Zhang, Wang, Zhang, Li, Xiong, Shi, Huang, Fu, and Zheng}]{zhang2024chemicaltextmining}
Zhang, W.; Wang, Q.; Kong, X.; Xiong, J.; Ni, S.; Cao, D.; Niu, B.; Chen, M.; Li, Y.; Zhang, R.; Wang, Y.; Zhang, L.; Li, X.; Xiong, Z.; Shi, Q.; Huang, Z.; Fu, Z.; and Zheng, M. 2024.
\newblock Fine-Tuning Large Language Models for Chemical Text Mining.
\newblock \emph{Chemical Science}, 15(27): 10600--10611.

\bibitem[{Zhao et~al.(2025{\natexlab{a}})Zhao, Brabson, Chheda, Huang, Kim, Liu, Mochida, Pham, Prerna, Terrones, Yoon, Zoubritzky, Coudert, Haranczyk, Kulik, Moosavi, Sholl, Siepmann, Snurr, and Chung}]{zhao2025coremofdb}
Zhao, G.; Brabson, L.; Chheda, S.; Huang, J.; Kim, H.; Liu, K.; Mochida, K.; Pham, T.; Prerna; Terrones, G.; Yoon, S.; Zoubritzky, L.; Coudert, F.-X.; Haranczyk, M.; Kulik, H.~J.; Moosavi, S.~M.; Sholl, D.~S.; Siepmann, J.~I.; Snurr, R.~Q.; and Chung, Y.~G. 2025{\natexlab{a}}.
\newblock {CoRE MOF DB}: A curated experimental metal--organic framework database with machine-learned properties for integrated material-process screening.
\newblock \emph{Matter}, 8(6): 102140.

\bibitem[{Zhao, Zhao, and Chung(2025)}]{zhao2025mofclassifier}
Zhao, G.; Zhao, P.; and Chung, Y.~G. 2025.
\newblock {MOFClassifier}: A Machine Learning Approach for Validating Computation-Ready Metal--Organic Frameworks.
\newblock \emph{Journal of the American Chemical Society}, 147(37): 33343--33349.

\bibitem[{Zhao et~al.(2025{\natexlab{b}})Zhao, Ma, Chen, Sun, Li, Xia, Chen, Xu, Zhu, Zhu, Fan, Shen, Yu, and Chen}]{zhao2024chemdfm}
Zhao, Z.; Ma, D.; Chen, L.; Sun, L.; Li, Z.; Xia, Y.; Chen, B.; Xu, H.; Zhu, Z.; Zhu, S.; Fan, S.; Shen, G.; Yu, K.; and Chen, X. 2025{\natexlab{b}}.
\newblock Developing {ChemDFM} as a Large Language Foundation Model for Chemistry.
\newblock \emph{Cell Reports Physical Science}, 6(4): 102523.

\bibitem[{Zhou et~al.(2024)Zhou, Yan, Shlapentokh-Rothman, Wang, and Wang}]{zhou2024lats}
Zhou, A.; Yan, K.; Shlapentokh-Rothman, M.; Wang, H.; and Wang, Y.-X. 2024.
\newblock Language Agent Tree Search Unifies Reasoning, Acting, and Planning in Language Models.
\newblock In \emph{Proceedings of the 41st International Conference on Machine Learning}, volume 235 of \emph{Proceedings of Machine Learning Research}, 62138--62160. PMLR.

\bibitem[{Zhuge et~al.(2024)Zhuge, Wang, Kirsch, Faccio, Khizbullin, and Schmidhuber}]{zhuge2024gptswarm}
Zhuge, M.; Wang, W.; Kirsch, L.; Faccio, F.; Khizbullin, D.; and Schmidhuber, J. 2024.
\newblock {GPTSwarm}: Language Agents as Optimizable Graphs.
\newblock In \emph{Proceedings of the 41st International Conference on Machine Learning}, volume 235 of \emph{Proceedings of Machine Learning Research}, 62743--62767. PMLR.

\end{thebibliography}
\appendix
\raggedbottom
\setcounter{secnumdepth}{2}   
\twocolumn
\begin{tcolorbox}[
  enhanced,
  colback=gray!2,
  colframe=rulegray!55,
  boxrule=0.5pt,
  arc=2pt,
  outer arc=2pt,
  left=5pt,
  right=5pt,
  top=4pt,
  bottom=4pt,
  before skip=0pt,
  after skip=6pt
]
\centering
{\small\bfseries Appendix Guide}\par
\vspace{3pt}
{\color{rulegray!45}\hrule}
\vspace{3pt}
\footnotesize
\renewcommand{\arraystretch}{1.08}
\begin{tabular*}{\linewidth}{@{}p{\linewidth}@{}}
{\footnotesize\bfseries\color{rulegray}SECTION} \\
\textbf{A}\enspace Extended Methodology \\
{\footnotesize\hspace{1.15em}A.1\enspace Fine-Grained Diagnostic Vocabulary} \\
{\footnotesize\hspace{1.15em}A.2\enspace GRPO Objective and Exact Reward Specification} \\
{\footnotesize\hspace{1.15em}A.3\enspace Frozen Two-Pass Inference Protocol} \\
{\footnotesize\hspace{1.15em}A.4\enspace Forensic Lab and Evidence Report} \\
{\footnotesize\hspace{1.15em}A.5\enspace Components of Chemically Grounded Diagnosis} \\

\textbf{B}\enspace Case Study \\

\textbf{C}\enspace Additional Experiments and Analysis \\
{\footnotesize\hspace{1.15em}C.1\enspace Evaluation Datasets} \\
{\footnotesize\hspace{1.15em}C.2\enspace Structural Distribution Shift} \\
{\footnotesize\hspace{1.15em}C.3\enspace Implementation Details} \\
{\footnotesize\hspace{1.15em}C.4\enspace Accuracy Is Not Explanation Quality} \\
{\footnotesize\hspace{1.15em}C.5\enspace Detailed Error Analysis by Error Family} \\

\textbf{D}\enspace Human Expert Validation \\
{\footnotesize\hspace{1.15em}D.1\enspace Protocol} \\
{\footnotesize\hspace{1.15em}D.2\enspace Attribution Findings} \\

\textbf{E}\enspace Interactive Audit Demo \\
\end{tabular*}
\end{tcolorbox}

\appendixanchor{appendix-extended-methodology}
\section{Extended Methodology}
\label{app:extended-methodology}

\appendixanchor{appendix-error-taxonomy}
\subsection{Fine-Grained Diagnostic Vocabulary}
\label{app:error-taxonomy}

\noindent This subsection documents the fixed 15-type diagnostic vocabulary
emitted by Sleuth. The labels are multi-label attribution outputs rather than
mutually exclusive classes. As defined in the Problem Definition, the large-scale
training labels and main benchmark metrics use four annotated parent families
through the mapping $m$, while fine-label expert analyses evaluate the 15-type
attribution directly when such annotations are available. Fixing this vocabulary
also lets the parser, reward verifiers, and Chem-GD apply the same deterministic
label constraints.

\begin{table*}[t]
\centering
\footnotesize
\setlength{\tabcolsep}{3.5pt}
\begin{tabular}{p{0.24\textwidth}p{0.16\textwidth}p{0.50\textwidth}}
\toprule
\rowcolor{groupbg}
\textbf{Fine Error Type} & \textbf{Parent $m(y)$} & \textbf{Description} \\
\midrule
\textbf{missing hydrogen} &
hydrogen &
Hydrogen atoms are missing from chemically plausible sites. \\
\textbf{wrong number of protons} &
hydrogen, charge &
The protonation state is inconsistent with the local chemistry or charge. \\
\textbf{terminal oxygen atom} &
hydrogen, charge &
Terminal oxygen suggests missing H, unresolved protonation, or charge imbalance. \\
\textbf{missing cations} &
charge &
Charge-balancing cations are absent or underrepresented. \\
\textbf{missing anions} &
charge &
Charge-balancing anions are absent or underrepresented. \\
\textbf{ion disorder} &
disorder &
Ionic species are disordered or ambiguously placed. \\
\textbf{occupancy} &
disorder &
Atoms exhibit abnormal or partial occupancy patterns. \\
\textbf{atoms overlapping} &
disorder &
Atoms are placed unrealistically close to one another. \\
\textbf{inverted conformers} &
disorder &
Local conformers or linker orientations are geometrically inconsistent. \\
\textbf{missing ligand} &
other &
A ligand or substantial linker fragment is missing. \\
\textbf{has over-coordinated atoms} &
other &
Atoms exceed chemically plausible coordination environments. \\
\textbf{has under-coordinated atoms} &
other &
Atoms have incomplete or chemically implausible coordination environments. \\
\textbf{without metal} &
other &
The structure lacks metal atoms expected for a MOF. \\
\textbf{without carbon} &
other &
The structure lacks carbon atoms expected for organic linkers. \\
\textbf{other} &
other &
Other chemically meaningful CIF errors not covered above. \\
\bottomrule
\end{tabular}
\caption{The 15 structured audit error types and their fixed mapping to the
four parent type families used by $R_{\mathrm{type}}$ and Type-Hit Acc. Free
solvent is context only and maps to no effective type.}
\label{tab:error-taxonomy}
\end{table*}
\appendixanchor{appendix-reward-specification}
\subsection{GRPO Objective and Exact Reward Specification}
\label{app:reward-specification}

\noindent\textbf{Group-relative normalization.}
For each report $r_i$, the policy $\pi_\theta$ samples $G$ audits
$\{o_{i,g}\}_{g=1}^{G}$. Using the utility in
Eq.~\ref{eq:structured-audit-utility}, GRPO normalizes their rewards within the
same report:
\begin{equation}
    \mu_i=\frac{1}{G}\sum_{g=1}^{G}R_{i,g}, \qquad
    \sigma_i=\sqrt{\frac{1}{G}\sum_{g=1}^{G}(R_{i,g}-\mu_i)^2},
\end{equation}
where $\mu_i$ and $\sigma_i$ are the within-group mean and standard deviation.
The advantage is
\begin{equation}
    \hat{A}_{i,g}=\frac{R_{i,g}-\mu_i}{\sigma_i+\epsilon_A},
\end{equation}
where $\epsilon_A>0$ is a numerical-stability constant; positive advantages
indicate audits that outperform alternatives for the same CIF.

\noindent\textbf{Policy update.}
The policy ratio is
\begin{equation}
    \rho_{i,g}(\theta)=
    \frac{\pi_\theta(o_{i,g}|r_i)}
    {\pi_{\theta_{\mathrm{old}}}(o_{i,g}|r_i)},
\end{equation}
where $\pi_{\theta_{\mathrm{old}}}$ is the policy before the current update. We
use the clipped surrogate
\begin{equation}
\begin{aligned}
L^{\mathrm{clip}}_{i,g}(\theta)
=\min\{&
\rho_{i,g}(\theta)\hat{A}_{i,g},\\
&\mathrm{clip}(\rho_{i,g}(\theta),1-\epsilon,1+\epsilon)
\hat{A}_{i,g}\},
\end{aligned}
\end{equation}
where $\epsilon$ is the clipping radius, and maximize
\begin{equation}
\begin{aligned}
\mathcal{J}(\theta)
=\mathbb{E}_{i,g}\big[
&L^{\mathrm{clip}}_{i,g}(\theta)\\
&-\beta D_{\mathrm{KL}}\big(
\pi_\theta(\cdot|r_i)\,\|\,\pi_{\mathrm{ref}}(\cdot|r_i)
\big)\big],
\end{aligned}
\end{equation}
where $\pi_{\mathrm{ref}}$ is a fixed reference policy, $D_{\mathrm{KL}}$ is the
Kullback--Leibler divergence, $\beta$ controls the strength of the
reference-policy penalty, and $\mathbb{E}_{i,g}$ averages over examples and
within-report samples. The KL term discourages unstable drift from the
reference auditor unless the structured audit reward supports the change.

\noindent\textbf{Exact verifier rewards.}
Table~\ref{tab:reward-specification} gives the exact per-completion rules and
selected weights. Let
$\hat{\mathcal{P}}=m(\hat{\mathcal{Y}})$ be the predicted parent-family set.
A predicted fine label is \emph{effective} when it maps to at least one parent
family. For grounding, let
$b_{\mathrm{grd}}=p_{\mathrm{cite}}\min(1,n_{\mathrm{field}}/2)$, where
$p_{\mathrm{cite}}$ is the precision of extracted field--value citations
against the report and $n_{\mathrm{field}}$ is the number of distinct cited
field paths.

\begin{table*}[t]
\centering
\scriptsize
\setlength{\tabcolsep}{4pt}
\begin{tabular}{p{0.115\textwidth}p{0.735\textwidth}c}
\toprule
Component & Exact per-completion scoring rule & Weight \\
\midrule
$R_{\mathrm{fmt}}$ &
$-1$ if the answer is unparseable or the fine-label list is invalid; $0$ if
the explanation is absent, has fewer than 30 characters, or uses the deprecated
reasoning protocol; $1$ otherwise. & $0.05$ \\
$R_{\mathrm{ans}}$ &
$1$ for a correct verdict, $-1$ for an unparseable verdict, $-0.85$ for a
false positive, and $-0.25$ for a false negative. & $2.00$ \\
$R_{\mathrm{type}}$ &
$0$ for an invalid fine-label list. If $\mathcal{P}^{\star}=\emptyset$, the
score is $1$ when $\hat{\mathcal{P}}=\emptyset$ and $-0.85$ otherwise. If
$\mathcal{P}^{\star}\ne\emptyset$, the score is $-0.25$ when
$\hat{\mathcal{P}}=\emptyset$, $-0.35$ when the sets do not overlap, and their
set F1 otherwise. & $1.25$ \\
$R_{\mathrm{cons}}$ &
$0$ if the answer or fine-label list is invalid; otherwise $1$ when
$\hat a=1\Leftrightarrow\hat{\mathcal{P}}\ne\emptyset$ and $-1$ when this
condition is violated. & $0.05$ \\
$R_{\mathrm{grd}}$ &
$0$ when $b_{\mathrm{grd}}=0$ or parsing fails;
$b_{\mathrm{grd}}$ for a correct verdict; $0.25b_{\mathrm{grd}}$ for an
incorrect verdict whose predicted and reference families overlap; and
$-0.2b_{\mathrm{grd}}$ otherwise. & $0.15$ \\
$R_{\mathrm{evid}}$ &
$0$ for an invalid fine-label list. With no effective predicted label, the
score is $0.5$ if the report has no hard flag and $-0.5$ otherwise. With
effective labels, the score is $1$ if all pass their label-specific support
rules, $0.25$ if only some pass, and $-1$ if none pass. & $0.90$ \\
\bottomrule
\end{tabular}
\caption{Exact reward rules and selected weights. Citation values match
numeric report values within $10^{-4}$ and nonnumeric values after
case-normalized exact matching. Label support is evaluated against frozen
predicates over report facts, diagnostic signals, and hard flags.}
\label{tab:reward-specification}
\end{table*}

\appendixanchor{appendix-frozen-inference-protocol}
\subsection{Frozen Two-Pass Inference Protocol}
\label{app:frozen-inference-protocol}

\begin{algorithm}[t]
\caption{Frozen conditional attribution with an immutable verdict}
\label{alg:frozen-inference}
\footnotesize
\begin{algorithmic}[1]
\REQUIRE CIF $x$, $G_0$, $G_+$, $f_\theta$, $I_0$, $I_+$, $\Pi$, and $M$
\ENSURE Final structured audit $o$
\STATE $r_0 \gets G_0(x)$
\STATE $o_0 \gets f_\theta(I_0,r_0)$
\STATE $(z_0,\mathcal{Y}_0,a_0) \gets \Pi(o_0)$
\IF{$a_0=0$}
    \RETURN $o_0$ \COMMENT{retain the clean first-pass audit unchanged}
\ENDIF
\STATE $r_+ \gets G_+(x,r_0)$
\STATE $o_+ \gets f_\theta(I_+,r_+)$
\STATE $(z_+,\mathcal{Y}_+,\_) \gets \Pi(o_+)$ \COMMENT{discard the second-pass verdict}
\IF{$z_+=\emptyset$}
    \STATE $z_+ \gets z_0$
\ENDIF
\IF{$\mathcal{Y}_+=\emptyset$}
    \STATE $\mathcal{Y}_+ \gets \mathcal{Y}_0$
\ENDIF
\STATE $H \gets \operatorname{hard}(r_+)$
\STATE $\mathcal{Y} \gets \operatorname{canonicalize}
       \bigl(\mathcal{Y}_+ \cup \operatorname{labels}(H)\bigr)$
\STATE $z \gets \operatorname{appendEvidence}(z_+,H)$
\RETURN $M(z,\mathcal{Y},a_0)$ \COMMENT{the final verdict remains $a_0=1$}
\end{algorithmic}
\end{algorithm}

\noindent A deterministic parser extracts the first complete
\texttt{<reasoning>}, \texttt{<error\_type>}, and \texttt{<answer>} blocks,
then normalizes only exact vocabulary labels. A method may use its declared,
bounded same-model retry for a parse failure; any field still malformed or
missing in the final stored output is counted as a failure. The merger copies
the first-pass verdict. If that verdict is clean, the complete first-pass audit
is retained; otherwise, the second-pass explanation and labels replace the
initial attribution, after which every deterministic hard-flag label and its
exact field--value citation are inserted. Assertions verify verdict identity
for every merged example.

\noindent Algorithm~\ref{alg:frozen-inference} specifies the deployed
protocol after GRPO training; the auditor parameters remain frozen. Let $G_0$
denote the base-report renderer, $G_+$ the enriched forensic-report renderer,
$f_\theta$ the frozen auditor, $I_0$ the base instruction, $I_+$ the attribution
instruction, $\Pi$ the deterministic XML parser, and $M$ the
deterministic schema merger. For a report $r$, $\operatorname{hard}(r)$ returns
the $\mathcal{H}$ component of the report, i.e., typed hard-flag
label--citation pairs, and $\operatorname{labels}$ extracts their labels.
$\operatorname{canonicalize}$ keeps only valid labels in the fixed output
order; $\operatorname{appendEvidence}$ adds only missing hard-flag citations to
an explanation.

\appendixanchor{appendix-forensic-lab}
\subsection{Forensic Lab and Evidence Report}
\label{app:forensic-lab}

\noindent\textbf{The Forensic Lab is a measurement layer, not another
decision maker.}
Its role is to turn a dense CIF into a compact evidence report before any
language-model reasoning. The tools read only parsed CIF content---atom types,
coordinates, occupancies, cell parameters, and periodic geometry---and never
use filenames, dataset labels, human notes, or model outputs. We distinguish
fully objective measurements, such as composition and interatomic distances,
from heuristic chemical summaries, such as functional-group assignment and
formal charge ledgers. This distinction is important: the Forensic Lab exposes
evidence and uncertainty, while the Sleuth agent decides whether the evidence
supports an audit label.

\begin{table*}[ht!]
\centering
\footnotesize
\setlength{\tabcolsep}{4pt}
\renewcommand{\arraystretch}{1.12}
\begin{tabular}{p{0.27\textwidth}p{0.67\textwidth}}
\toprule
\rowcolor{groupbg}
\textbf{Forensic tool} & \textbf{Evidence produced for the auditor} \\
\midrule
\textbf{CIF reader} &
Parses atom sites, element symbols, coordinates, occupancies, periodic cell
parameters, and formula-level metadata. \\
\textbf{Composition--cell checker} &
Reports formula, atom count, element counts, \zH/\zC ratio, lattice lengths, angles,
and cell volume as objective facts. \\
\textbf{Occupancy checker} &
Detects partial occupancies and split-site hints, exposing direct evidence for
disorder or unresolved crystallographic sites. \\
\textbf{Geometry checker} &
Computes PBC neighbor distances, minimum distances, heavy-atom overlaps, and
short \zH--X contacts; this acts like a ruler for impossible local geometry. \\
\textbf{Cutoff-graph analyzer} &
Builds a distance-cutoff connectivity graph, then reports connected components,
small non-metal fragments, possible counter-ions, and unidentified fragments. \\
\textbf{Functional-group classifier} &
Identifies chemically relevant motifs such as aromatic \zC lacking \zH,
carboxylate/protonated carboxylate, phosphonate, sulfonate, \zO/\zN environments,
and halides. \\
\textbf{Coordination--bridge analyzer} &
Summarizes metal coordination shells, over/under-coordination cues, bridging
\zO/halide/S atoms, and terminal-oxo warnings. \\
\textbf{Charge-ledger checker} &
Balances common metal oxidation states against anionic ligand contributions,
returning positive charge, negative charge, net charge, assumptions, and
uncertainty. \\
\textbf{Evidence compiler} &
Converts all measurements into candidate signals, co-occurrence signals, known
failure modes, and machine-readable aliases for citation checking. \\
\bottomrule
\end{tabular}
\caption{Forensic Lab components. The first eight rows are CIF-derived
checkers; the final row compiles their outputs into the report read by the
Sleuth agent.}
\label{tab:forensic-lab-tools}
\end{table*}

\noindent\textbf{The evidence report is a lab notebook that the model can
cite.}
After the checkers run, their outputs are rendered through a fixed report
interface used in training and evaluation: \emph{Machine-readable field
aliases}, \emph{Objective facts}, \emph{Hard flags}, \emph{Possible error
signals}, \emph{Co-occurring possible signals}, and \emph{Context for
interpretation}. Objective facts expose deterministic measurements; hard flags
are deterministic label--citation pairs; possible signals remain weighted
hypotheses rather than verdicts, and context records assumptions, uncertainty,
and known false-positive modes.
For conditional attribution, an additive forensic layer inserts hydrogen-host
counts, heavy-atom contacts, coordination histograms, and fragmentation signals
into these native sections without altering the base report. The auditor
receives the corresponding plain-text report serialization. For legibility, the example
below is a content-equivalent XML rendering of one held-out anionic framework
($\mathrm{\textcolor{elemC}{C}_{40}\textcolor{elemH!60!black}{H}_{16}\textcolor{elemZn}{Ag}_2\textcolor{elemN}{N}_8S_8}$);
the XML changes presentation only, not the fields, values, or model input.

\begin{figure}[t!]
\centering
\begin{minipage}{0.96\columnwidth}
\begin{tcblisting}{
  enhanced,
  listing only,
  listing engine=listings,
  listing options={
    style=evidencexml,
    frame=none,
    backgroundcolor=\color{gray!3},
    aboveskip=0pt,
    belowskip=0pt
  },
  title={FORENSIC EVIDENCE REPORT},
  fonttitle=\comicneue\small\bfseries,
  coltitle=toolblue,
  colbacktitle=toolblue!10,
  colback=gray!3,
  colframe=evidline,
  boxrule=0.45pt,
  arc=8pt,
  outer arc=8pt,
  titlerule=0.45pt,
  halign title=center,
  toptitle=4pt,
  bottomtitle=4pt,
  boxsep=0pt,
  left=5pt,
  right=5pt,
  top=4pt,
  bottom=4pt,
  before skip=0pt,
  after skip=0pt
}
<forensic_report>
  <field_aliases>
    <field name="composition.formula"
           value="C40H16Ag2N8S8"/>
    <field name="geometry.min_distance_A"
           value="0.949"/>
    <field name="graph.n_components" value="1"/>
    <field name="metal_coordination.Ag.avg_N"
           value="4.0"/>
    <field name="charge.net" value="-6.0"/>
  </field_aliases>

  <objective_facts>
    <composition atom_count="74" H_to_C="0.4">
      <elements Ag="2" C="40" H="16" N="8" S="8"/>
    </composition>
    <cell a="13.638" b="13.638" c="6.205"
          alpha="90.0" beta="90.0" gamma="90.0"
          volume_A3="1154.1"/>
    <geometry min_distance_A="0.949"
              partial_occupancy_count="0"
              heavy_overlap_count="0"
              short_HX_contact_count="0"/>
    <graph components="1"
           largest_component_atoms="74"
           small_nonmetal_components="0"/>
    <functional_groups aromatic_C="16"
                       aromatic_C_lacking_H="0"
                       azolate_like_metal_bound_N="8"/>
    <metal_coordination element="Ag" sites="2"
                        avg_N="4.0" avg_O="0.0">
      <site id="Ag72" shell="4 N at 2.29 A"/>
      <site id="Ag73" shell="4 N at 2.29 A"/>
    </metal_coordination>
    <bridging_atoms element="S" mu0="8"/>
  </objective_facts>

  <hard_error_flags count="0"/>

  <!-- Heuristic candidate; not a final verdict. -->
  <possible_error_signal id="E1"
                         family="charge_or_h"
                         weight="medium">
    <evidence field="charge.net" value="-6.0"/>
    <rationale>
      Moderate net formal charge under common oxidation-state
      assumptions.
    </rationale>
    <candidate_interpretations>
      missing H | missing counter-ions | wrong oxidation states
    </candidate_interpretations>
  </possible_error_signal>

  <cooccurring_possible_signals count="0"/>

  <interpretation_context>
    <charge_ledger confidence="low">
      <positive source="Ag(+1) x 2">2</positive>
      <negative source="azolate-like N(-1) x 8">8.0</negative>
      <net>-6.0</net>
    </charge_ledger>
    <uncertainty>
      N-donor classification may confuse neutral pyridyl and
      anionic azolate sites.
    </uncertainty>
    <counter_ion_candidates count="0"/>
    <charge_correction target_abs_net_le="1.0"
                       positive_correction_needed="about 6"
                       candidates="none"/>
    <known_false_positive_contexts>
      mixed-valent metals | polyoxometalates | cationic
      frameworks with removed counter-anions | rare oxidation
      states | H-rider placement artifacts
    </known_false_positive_contexts>
  </interpretation_context>
</forensic_report>
\end{tcblisting}
\end{minipage}
\end{figure}

\noindent\textbf{The report grounds each explanation in checkable chemical
evidence.}
This representation gives the auditor a small set of chemically meaningful
handles instead of thousands of raw CIF tokens. If the Sleuth agent writes that
\texttt{charge.net=-6.0} supports a missing-cation diagnosis, the parser can
verify both parts of the claim: the cited field and value must exist in the
machine-readable aliases or report fields, and that cited field must be relevant
to the predicted type. The base report powers GRPO and first-pass inference;
the enriched report supplies the additional evidence used by conditional
attribution and final Chem-GD evaluation.

\newcommand{\rendercasestudyfigure}[1][t!]{%
\begin{figure*}[#1]
\centering
\begingroup
\comicneue
\noindent\begin{tikzpicture}
\node[draw=rulegray!75, line width=0.9pt, rounded corners=7pt, fill=white,
      inner sep=7pt, text width=\dimexpr\textwidth-17pt\relax]{%
\begin{tikzpicture}
\node[fill=groupbg, draw=rulegray!50, line width=0.6pt, rounded corners=5pt,
      inner sep=5pt, text width=\dimexpr\linewidth-11pt\relax]{%
\small
\begin{tabular*}{\linewidth}{@{\extracolsep{\fill}}lll@{}}
\textcolor{rulegray}{\scriptsize\bfseries CASE}\enspace
  \texttt{tobmof-3359} &
\textcolor{rulegray}{\scriptsize\bfseries STRUCTURE}\enspace
  $\mathrm{Zr_6O_4(OH)_4}$ node $+$ 12 carboxylate linkers &
\textcolor{rulegray}{\scriptsize\bfseries REFERENCE}\enspace
  \textcolor{okgreen}{\textbf{clean}}~\cmk
\end{tabular*}};
\end{tikzpicture}
\par
\vspace{5pt}
\noindent\begin{minipage}[t]{0.475\linewidth}
\vspace{0pt}
\casecrayonrule{badred}
\vspace{2pt}
\textbf{GPT-5.5}\hfill
\casebadge{badred}{$\times$\ FALSE POSITIVE}
\par
{\scriptsize\color{rulegray}Raw-CIF end-to-end audit}\par
\vspace{5pt}

\casecrayonlabel{badred}{INPUT}\par
\noindent\begin{tikzpicture}
\node[fill=gray!6, draw=gray!45, line width=0.7pt, rounded corners=4pt,
      inner sep=5pt, text width=\dimexpr\linewidth-11pt\relax]{%
{\fontsize{6.5}{8.1}\selectfont\ttfamily
C1~~C~~0.4701~~1.0011~-0.5557\\
C2~~C~~0.5344~~0.9901~-0.5577\\
H7~~H~~0.3927~~1.0026~-0.4957\\
\textcolor{rulegray}{...~~366 rows, no bonds}}
\par\vspace{3pt}
\centering
{\footnotesize\color{rulegray}substituted ring C}\par
\vspace{2pt}
{\large$\Downarrow$}\par
\vspace{1pt}
{\footnotesize\bfseries\color{badred}inferred ``missing H''}
};
\end{tikzpicture}
\par
\vspace{5pt}

\casecrayonlabel{badred}{EXPLANATION EXCERPT}\par
{\scriptsize\itshape ``C1--C6 form an \textbf{isolated benzene-like ring},
but only H7 and H8 are present; unsubstituted ring carbons $\dots$ have no
attached substituent atoms $\dots$ \textbf{indicating missing hydrogens}.''}
\par
\vspace{5pt}

\casecrayonlabel{badred}{STRUCTURED AUDIT}\par
\noindent\begin{tikzpicture}
\node[fill=badred!8, draw=badred!45, line width=0.7pt, rounded corners=3.5pt,
      inner sep=4pt, text width=\dimexpr\linewidth-9pt\relax]{%
\ttfamily\scriptsize
<error\_type> \textcolor{badred}{\textbf{["missing hydrogen"]}}
\hfill <answer> \textcolor{badred}{\textbf{1}}};
\end{tikzpicture}
\par
\vspace{3pt}
{\scriptsize\textcolor{badred}{\textbf{Failure:}} linker substitution is
mistaken for hydrogen deficiency.}
\end{minipage}\hfill
\begin{minipage}[t]{0.475\linewidth}
\vspace{0pt}
\casecrayonrule{okgreen}
\vspace{2pt}
\textbf{MOF-Sleuth}\hfill
\casebadge{okgreen}{\checkmark\ CORRECT}
\par
{\scriptsize\color{rulegray}Tool-grounded, RL-aligned audit}\par
\vspace{5pt}

\casecrayonlabel{okgreen}{FORENSIC EVIDENCE}\par
\noindent\begin{tikzpicture}
\node[fill=gray!6, draw=gray!45, line width=0.7pt, rounded corners=4pt,
      inner sep=5pt, text width=\dimexpr\linewidth-11pt\relax]{%
{\fontsize{6.5}{8.1}\selectfont\ttfamily
charge.net~=~0.0\hfill carboxylate.deprot~=~12\\
mu\_OH~/~mu\_oxo~=~4/4\hfill aromatic\_C.lackingH~=~0}\par
\vspace{4pt}
\centering
{\footnotesize
$\underbrace{12\,\mathrm{COO^-}+4\,\mu\text{-}\mathrm{OH^-}
+4\,\mu\text{-}\mathrm{O^{2-}}}_{24^-}
\;\textcolor{okgreen}{\boldsymbol{=}}\;
\underbrace{6\,\textcolor{elemZn}{\mathbf{Zr}}^{4+}}_{24^+}$}};
\end{tikzpicture}
\par
\vspace{5pt}

\casecrayonlabel{okgreen}{EXPLANATION EXCERPT}\par
{\scriptsize\itshape ``$\dots$deprotonated\_carboxylate=12.0, mu\_OH=4.0,
mu\_oxo\_x2=8.0 \textbf{sums to 24.0, matching positive=24.0} $\dots$
\textbf{no missing hydrogens}.''}
\par
\vspace{5pt}

\casecrayonlabel{okgreen}{STRUCTURED AUDIT}\par
\noindent\begin{tikzpicture}
\node[fill=okgreen!8, draw=okgreen!45, line width=0.7pt, rounded corners=3.5pt,
      inner sep=4pt, text width=\dimexpr\linewidth-9pt\relax]{%
\ttfamily\scriptsize
<error\_type> \textcolor{okgreen}{\textbf{[]}}
\hfill <answer> \textcolor{okgreen}{\textbf{0}}};
\end{tikzpicture}
\par
\vspace{3pt}
{\scriptsize\textcolor{okgreen}{\textbf{Verified:}} charge is balanced and no
aromatic carbon lacks hydrogen.}
\end{minipage}};
\end{tikzpicture}
\endgroup
\caption{Case study on clean ToBaCCo framework \texttt{tobmof-3359}. GPT-5.5
infers missing hydrogen from raw atom-site rows, whereas \textsc{MOF-Sleuth}
verifies charge balance and aromatic hydrogen occupancy before returning the
correct clean verdict. Explanation excerpts are verbatim.}
\label{fig:case}
\end{figure*}}

\appendixanchor{appendix-cgd-components}
\subsection{Components of Chemically Grounded Diagnosis}
\label{app:cgd-components}

Chemically Grounded Diagnosis combines verdict correctness with the evidence,
diagnosis, and evidence--label linkage predicates defined in the main text.
All components use the recognized diagnostic set returned by the parser;
context-only \emph{has free solvents} mentions and out-of-vocabulary strings
do not enter this set, while parseability is reported separately.
For diagnostic decomposition, Structural Evidence Fidelity (SEF) averages the
evidence predicate over all $N$ examples:
\begin{equation}
\mathrm{SEF}=\frac{1}{N}\sum_{i=1}^{N}g_i.
\end{equation}
Chemical Diagnosis Support (CDS) is label-level precision over attributable
links from verified cited evidence to predicted error types:
\begin{equation}
\mathrm{CDS}=
\frac{\sum_i\sum_{y\in\hat{\mathcal{Y}}_i}
\mathbb{I}[\exists q\in\mathcal{Q}(z_i):q\in\mathcal{B}_i
\wedge q\rightsquigarrow_i y]}
{\sum_i|\hat{\mathcal{Y}}_i|}.
\end{equation}
For report-based outputs, $q\rightsquigarrow_i y$ holds only when the cited
field--value pair matches the canonical report and its normalized field path
matches a frozen relevant-field pattern for $y$. These patterns are transcribed
from the fields consulted by the label-specific report-support rules. For
raw-CIF outputs, the same explanation sentence must contain a verified CIF field
or atom identifier and an explicit label-specific cue. The discriminative
predicate $\Delta_i$ first applies the frozen report-support rule to every
predicted type. It then rejects diagnoses containing only \emph{missing
anions} or \emph{missing cations} when fragmentation, hostless hydrogen,
heavy-atom clash, or linker mismatch is present but no structural type is
predicted. These component metrics isolate failure sources without requiring a
correct verdict; the main text therefore reports the stricter, class-balanced
Chem-GD.

\appendixanchor{appendix-case-study}
\section{Case Study}
\label{app:case-study}

\noindent\textbf{Tool grounding plus RL turns weak signals into correct
verdicts.}
Figure~\ref{fig:case} contrasts GPT-5.5---the strongest external
baseline---reading the raw CIF end-to-end with the RL-trained tool-grounded
auditor on the clean ToBaCCo framework \texttt{tobmof-3359}, a
\textcolor{elemZn}{Zr}$_6$\zO$_4$(\zO\zH)$_4$ node with twelve carboxylate
linkers. Because the CIF exposes only atom-site rows, the end-to-end model
cannot reconstruct the bonding graph: it sees benzene-like rings carrying
``only H7 and H8,'' treats the four substituted ring carbons as unsubstituted,
and flags \emph{missing hydrogen}---a hallucinated defect on a clean structure.
The tool-grounded auditor instead receives an explicit charge ledger and
integrates it: twelve deprotonated carboxylates ($-12$), four $\mu$-\zO\zH
($-4$), and four $\mu$-\textcolor{elemO}{o}xo units ($-8$) supply the $24$
negative charges that balance the $+24$ from the six
\textcolor{elemZn}{Zr}$^{4+}$ centers, and the aromatic-hydrogen check reports
no ring carbon lacking H, so the sparse hydrogen count is expected substitution
chemistry rather than a defect. It returns an empty error list and answer~0.
This is the
behavior $R_{\mathrm{evid}}$ targets: a diagnosis is emitted only when
the surrounding structural facts support it, not whenever a scalar looks
unusual.

\noindent\textbf{The case also explains the ToBaCCo gain.}
ToBaCCo contains many large framework structures where weak local cues can be
benign once charge balance, coordination, and functional-group accounting are
considered jointly. The frozen auditor over-flags such cases, producing 355
false positives on ToBaCCo, whereas the RL-trained auditor reduces this to 33
while adding only ten missed errors (false negatives $14\!\to\!24$). The ToBaCCo
accuracy gain ($0.631\!\to\!0.943$) is therefore not a mere threshold shift; it
reflects a learned policy for combining competing tool signals into a final
audit decision.

\rendercasestudyfigure[t!]

\appendixanchor{appendix-experimental-setup}
\section{Additional Experiments and Analysis}
\label{app:experimental-setup}

\appendixanchor{appendix-eval-datasets}
\subsection{Evaluation Datasets}
\label{app:eval-datasets}

The training split contains 8,006 human-annotated CIFs, including 3,959
erroneous and 4,047 clean structures. Separate development and internal
held-out splits contain 1,000 and 1,002 examples, respectively. Evaluation uses
CoRE-MOF 2019~\cite{chung2019coremof2019,gibaldi2025setc}, CoRE-MOF
2026~\cite{zhao2025coremofdb}, QMOF~\cite{rosen2021qmof}, and
ToBaCCo~\cite{colon2017tobacco}. The suite composition and sampling rules are
reported below. Parent-family type annotations are available for the CoRE-MOF
2019 evaluation split and are used for Type-Hit Acc; the external suites are
evaluated with binary labels and deterministic explanation-grounding metrics.

\begin{figure*}[t!]
\centering
\includegraphics[width=\textwidth]{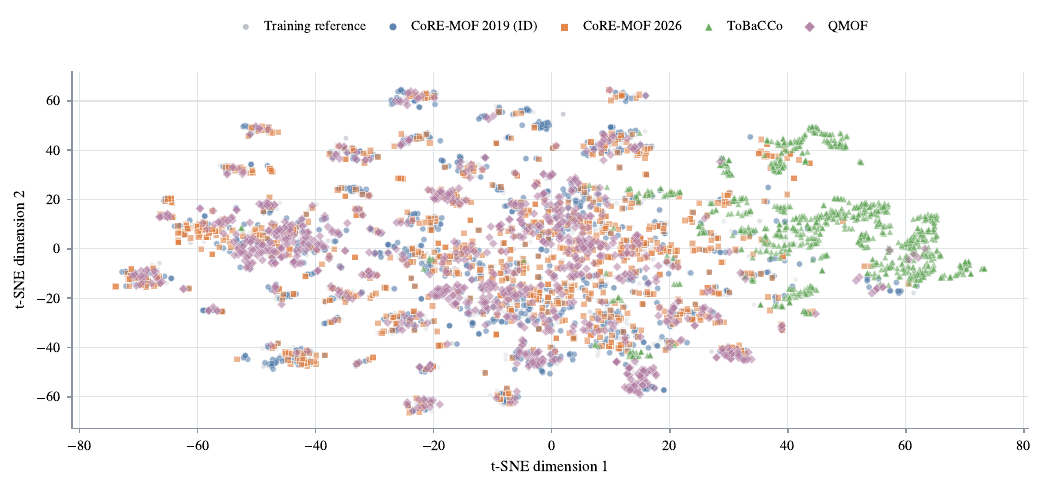}
\caption{Pooled t-SNE visualization of label-free structural descriptors for
the training reference and four evaluation suites. Gray points denote a
seed-matched sample of the training distribution. Blue points are the
CoRE-MOF 2019 held-out in-distribution test, whereas orange, green, and purple
points are the external CoRE-MOF 2026, ToBaCCo, and QMOF suites. All points use
the same preprocessing and one joint projection. The external suites exhibit
different structural coverage, most visibly the shifted ToBaCCo neighborhoods,
relative to the training and held-out CoRE-MOF reference. This visualizes the
external-suite diversity summarized in Table~\ref{tab:eval-datasets}.}
\label{fig:dataset-shift}
\end{figure*}

\begin{table*}[t!]
\centering
\small
\setlength{\tabcolsep}{6pt}
\begin{tabular}{p{0.15\textwidth}p{0.39\textwidth}rrrp{0.20\textwidth}}
\toprule
Dataset & Source / sampling rule & Total & Clean & Error & Purpose \\
\midrule
CoRE-MOF 2019~\cite{chung2019coremof2019,gibaldi2025setc} & seed-42 sample from the CoRE-MOF 2019 held-out split & 1000 & 481 & 519 & in-distribution test \\
CoRE-MOF 2026~\cite{zhao2025coremofdb} & 500 clean + 500 erroneous structures & 1000 & 500 & 500 & external balanced test \\
QMOF~\cite{rosen2021qmof} & all 23 erroneous structures + seed-42 clean fill & 1000 & 977 & 23 & imbalanced stress test \\
ToBaCCo~\cite{colon2017tobacco} & 500 clean + 500 erroneous structures & 1000 & 500 & 500 & external balanced test \\
\bottomrule
\end{tabular}
\caption{Evaluation datasets.}
\label{tab:eval-datasets}
\end{table*}

\appendixanchor{appendix-distribution-shift}
\subsection{Structural Distribution Shift}
\label{app:distribution-shift}

To inspect covariate shift independently of labels and audit outcomes, we pool
a seed-42 sample of 1,000 training structures with the four 1,000-structure
evaluation suites and embed them jointly in Figure~\ref{fig:dataset-shift}.
Each structure is represented only by 87 label-free descriptors comprising
elemental atomic fractions, atom count, unit-cell volume, an atom-density
proxy, normalized cell lengths, and cell angles. We standardize the pooled
descriptors, reduce them to 30 principal components, and fit one t-SNE
projection with seed 42 and perplexity 40. The analysis uses neither labels nor
audit outcomes. It characterizes structural coverage of the evaluation suites
and is not used for training, threshold selection, or metric computation.

We complement the visualization with an RBF maximum mean discrepancy
permutation test in the original descriptor space, standardized using only the
training reference. The held-out CoRE-MOF 2019 split shows no detectable shift
from training under this representation ($\mathrm{MMD}^2=0.0011$, $p=0.646$), whereas
CoRE-MOF 2026 ($0.0102$), ToBaCCo ($0.3455$), and QMOF ($0.0380$) are each
shifted with $p=0.002$ over 499 permutations. All three external comparisons
remain significant after Bonferroni correction for the three OOD tests. Thus,
the external suites are statistically shifted under label-free structural
descriptors while the held-out CoRE-MOF split is not.

\appendixanchor{appendix-implementation-details}
\subsection{Implementation Details}
\label{app:implementation-details}

\noindent\textbf{Baseline implementation.}
All applicable baselines use a shared task interface and evaluation protocol.
All LLM baselines use the same output contract and parser. Agent scaffolds share
the raw CIF input and frozen Qwen3-4B backbone, retain their method-specific
multi-call behavior, and never receive our tools or weight updates.
For MOF-specific validators, we evaluate released checkpoints on the ToBaCCo
benchmark used in Table~\ref{tab:ablation}. We retain the MOFClassifier core
ensemble~\cite{zhao2025mofclassifier} and SETC-GAT atomic
checkpoint~\cite{gibaldi2025setc}, and compare them with
MOFChecker~2.0~\cite{jin2025mofchecker}. ToBaCCo is a balanced external suite
and shows the largest label-free descriptor shift from our training reference
(Figure~\ref{fig:dataset-shift}); unsupported inputs count as incorrect.

\noindent\textbf{Training and inference configuration.}
Table~\ref{tab:experiment-config} summarizes the hardware, selection, GRPO, and
decoding settings used for the final auditor.

\begin{table}[t]
\centering
\footnotesize
\setlength{\tabcolsep}{3.5pt}
\begin{tabular}{>{\raggedright\arraybackslash}p{0.29\columnwidth}
                >{\raggedright\arraybackslash}p{0.62\columnwidth}}
\toprule
Item & Setting \\
\midrule
Hardware &
Internal cluster with 64 NVIDIA H100 GPUs; the selected GRPO job uses eight
H100 GPUs. \\
Model and tuning &
Qwen3-4B-Instruct, full-parameter tuning, Qwen no-thinking chat template. \\
Selection protocol &
Reward weights, hyperparameters, and checkpoints are selected only on a
CoRE-MOF 2019 development subset disjoint from all reported test sets. All
4,000 binary test labels are manually verified. \\
GRPO sampling &
Eight completions per report; four sample generations logged; temperature
1.1, top-$p$ 0.9, top-$k$ 20. \\
Optimization &
350 steps; learning rate $2\times10^{-6}$; cosine schedule; warmup ratio 0.1;
KL coefficient $\beta=0.04$; checkpoints every 50 steps; selected checkpoint
step 250. \\
Length and batch &
Maximum context length 8192; maximum completion length 1024; per-device batch
size 8; gradient accumulation 8. \\
Systems settings &
bfloat16 precision; DeepSpeed ZeRO-2; gradient checkpointing; colocated vLLM;
vLLM memory utilization 0.30; four dataloader workers; eight dataset workers;
seed 42. \\
Reward and inference &
Reward weights are listed in Table~\ref{tab:reward-specification}. Final
inference uses greedy decoding. Sleuth first generates a complete audit; for
predicted-error cases, the same LLM performs a verdict-preserving attribution
pass to reduce unsupported attributions and hallucinated explanations. A
deterministic merger keeps the original binary answer unchanged. \\
\bottomrule
\end{tabular}
\caption{Experimental configuration for the final \textsc{MOF-Sleuth} auditor.
Fine-grained expert attribution is evaluated separately in
Appendix D.}
\label{tab:experiment-config}
\end{table}

\appendixanchor{appendix-chemical-grounding}
\subsection{Chemically Grounded Diagnosis: Accuracy Is Not Explanation Quality}
\label{app:chemical-grounding}

\begin{figure}[t]
\centering
\includegraphics[width=0.82\columnwidth]{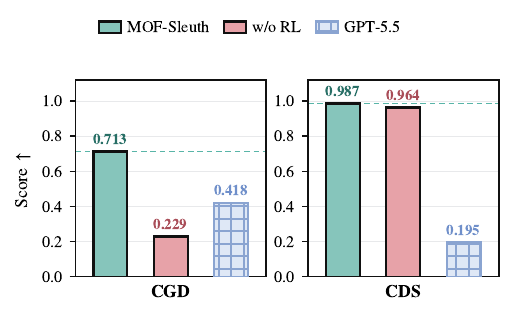}
\caption{Grounded-explanation quality for GPT-5.5 and \textsc{MOF-Sleuth}
with and without RL. Tool evidence yields high CDS without RL, while
reward-guided alignment raises the stricter Chem-GD score.}
\label{fig:chemical-grounding}
\end{figure}

\noindent A correct verdict is not enough for CIF curation: the explanation
must cite real chemical evidence and use it to support the diagnosis.
Figure~\ref{fig:chemical-grounding} contrasts GPT-5.5 with
\textsc{MOF-Sleuth} using the component scores defined in
Appendix A.5. GPT-5.5 reaches Chem-GD $0.418$ but only
CDS $0.195$, indicating weak diagnosis support. The frozen tool-report auditor
already reaches CDS $0.964$, yet its Chem-GD remains $0.229$; available
tool-supported labels alone do not ensure a correct complete audit.
Reward-guided alignment preserves CDS ($0.987$) while raising Chem-GD to
$0.713$.

Table~\ref{tab:cgd-components} provides the complete SEF/CDS decomposition for
all evaluated language-model and agent baselines.

\begin{table}[t]
\centering
\scriptsize
\setlength{\tabcolsep}{4pt}
\renewcommand{\arraystretch}{0.92}
\begin{tabular}{@{}p{0.64\columnwidth}cc@{}}
\toprule
Method & SEF\up & CDS\up \\
\midrule
\rowcolor{groupbg}
\multicolumn{3}{l}{\textit{End-to-end LLMs}} \\
Qwen3-4B~\cite{yang2025qwen3} & 0.527 & 0.071 \\
Qwen3-8B~\cite{yang2025qwen3} & 0.686 & 0.016 \\
Qwen3-30B-A3B~\cite{yang2025qwen3} & 0.786 & 0.056 \\
Gemma-4-31B~\cite{google2026gemma4} & 0.844 & 0.087 \\
DeepSeek-v4-Pro~\cite{deepseek2026v4pro} & 0.629 & 0.229 \\
GPT-5.5~\cite{openai2026gpt55} & 0.940 & 0.195 \\
Claude-Sonnet-4.6~\cite{anthropic2026sonnet46} & 0.670 & 0.366 \\
\midrule
\rowcolor{groupbg}
\multicolumn{3}{l}{\textit{Agent-scaffold pipelines}} \\
Self-Consistency~\cite{wang2023selfconsistency} & 0.523 & 0.067 \\
Reflexion~\cite{shinn2023reflexion} & 0.471 & 0.079 \\
Tree-of-Thoughts~\cite{yao2023tot} & 0.566 & 0.131 \\
LATS~\cite{zhou2024lats} & 0.531 & 0.088 \\
AutoGen~\cite{wu2024autogen} & 0.547 & 0.067 \\
CrewAI~\cite{crewai2024} & 0.502 & 0.103 \\
MetaGPT~\cite{hong2024metagpt} & 0.511 & 0.067 \\
LangGraph~\cite{langgraph2024} & 0.511 & 0.071 \\
DSPy~\cite{khattab2024dspy} & 0.525 & 0.145 \\
GPTSwarm~\cite{zhuge2024gptswarm} & 0.432 & 0.074 \\
AFlow~\cite{zhang2025aflow} & 0.646 & 0.093 \\
AgentSquare~\cite{shang2025agentsquare} & 0.672 & 0.081 \\
ADAS~\cite{hu2025adas} & 0.602 & 0.071 \\
\midrule
\rowcolor{groupbg}
\multicolumn{3}{l}{\textit{Ours}} \\
\rowcolor{oursbg}
\textsc{MOF-Sleuth} w/o RL & 0.621 & 0.964 \\
\rowcolor{oursbg}
\textbf{\textsc{MOF-Sleuth}} & \textbf{0.974} & \textbf{0.987} \\
\bottomrule
\end{tabular}
\caption{Complete component metrics underlying Chemically Grounded Diagnosis.
SEF is Structural Evidence Fidelity; CDS is Chemical Diagnosis Support.}
\label{tab:cgd-components}
\end{table}

\noindent\textbf{Development, validation, and scope.}
The discriminative support rule was specified on a deterministic hash-split
development half of the ToBaCCo expert study and frozen before scoring the
other half. Agreement between automatic pass/fail and expert attribution
correctness rose from 0.519 for broad report compatibility to 0.722 on this
held-out half. This is development evidence because both halves originate from
the same ToBaCCo study. After freezing the complete protocol, we applied it
without modification to an independent 100-case panel with 25 examples from
each evaluation suite. Final Chem-GD attains 0.720 agreement, 0.938 precision,
and 0.714 recall for identifying expert-correct attributions on this cross-suite
panel. Thus, the expert study provides initial external validation of Chem-GD as
a high-precision deterministic proxy for expert-judged audit success.

Chem-GD evaluates the complete tool-agent audit, including deterministic
postprocessing, rather than unaided free-form LLM reasoning. It shares the
frozen verifier family with training, but the roles are distinct:
$R_{\mathrm{grd}}$ trains evidence truthfulness, $R_{\mathrm{evid}}$ trains
diagnostic support, and Chem-GD tests the final correct audit for both.
Reward ablations that improve Chem-GD therefore demonstrate alignment to this
objective, while the blinded cross-suite panel checks metric meaningfulness
independently. Chem-GD does not claim exact recovery of every fine-grained
label or causal validity of every explanation statement, and it does not
replace expert chemical ground truth.

\appendixanchor{appendix-errtype}
\subsection{Detailed Error Analysis by Error Family}
\label{app:errtype}

\noindent Because only CoRE-MOF 2019~\cite{chung2019coremof2019,gibaldi2025setc}
provides human error-family annotations, we report per-family F1 to show where
typed diagnosis remains difficult beyond binary detection.

\begin{figure}[t]
\centering
\includegraphics[width=0.92\columnwidth]{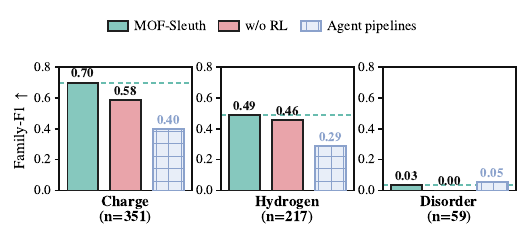}
\caption{Per-family attribution F1 on
CoRE-MOF 2019~\cite{chung2019coremof2019,gibaldi2025setc}.
\textsc{MOF-Sleuth} improves the dominant charge and hydrogen families;
``Agent pipelines'' averages the 13 raw-CIF agent baselines, and the rare
\emph{other} family is omitted.}
\label{fig:errtype}
\end{figure}

Remaining errors concentrate in overlapping charge, protonation, and
terminal-oxygen families, where several labels can be structurally plausible.
This supports evidence-grounded attribution rather than binary detection alone.

\appendixanchor{appendix-human-eval}
\section{Human Expert Validation}
\label{app:human-eval}

\noindent Expert CIF review is time intensive because each case requires
checking atom-site rows, charge/protonation, occupancy, connectivity, and
coordination evidence. We therefore use blinded expert panels with fixed seeds
to externally validate three claims: the verdicts are reliable, the fine-grained
causes are chemically useful, and verdict-preserving attribution refinement
improves explanations after the protocol is frozen. One stratified 100-case
ToBaCCo panel freezes the refinement protocol, and an
independent 100-case cross-suite panel tests transfer
(Table~\ref{tab:human-validation}).

\appendixanchor{appendix-human-protocol}
\subsection{Protocol}
\label{app:human-protocol}

\noindent We randomly sampled 100 ToBaCCo structures with seed 42, stratified
into 50 existing-error and 50 existing-clean cases. Two domain experts (Expert
A and Expert B) independently reviewed anonymized CIFs---without filename,
database label, tool report, or model output---and recorded a verdict,
fine-grained causes, confidence, and decisive atom/site evidence. Free solvent
alone remained context rather than an effective error. Disagreements were
adjudicated into a consensus set. Model outputs and binary verdicts were hidden
until adjudication, preventing annotation anchoring. This first panel froze the
verdict-preserving attribution-refinement protocol; the original verdicts
remained immutable. We then applied the frozen protocol unchanged to a
separately annotated panel containing 25 examples from each evaluation suite.

\appendixanchor{appendix-human-attribution}
\subsection{Attribution Findings}
\label{app:human-attribution}

\noindent\textbf{Blinded experts confirm reliable verdicts and useful causes.}
Table~\ref{tab:human-validation}(a) shows 0.930 verdict agreement and
$\kappa=0.860$ with consensus, with 47 TP, 46 TN, 4 FP, and 3 FN. The 15-label
attribution task is intentionally stringent: exact-set expert agreement is
54\%, and direct expert Jaccard/F1 is 0.252/0.389. Against the adjudicated
consensus, \textsc{MOF-Sleuth} reaches 0.700 Type-Hit Acc and 0.294/0.418
Jaccard/F1, placing its fine-grained suggestions within the observed
expert-disagreement range. This supports the 15-type vocabulary as an
actionable attribution interface for database curation.

\noindent\textbf{Conditional attribution improves explanations without changing
answers.}
On ToBaCCo-100, refinement raises Type-Hit Acc from 0.260 to 0.700, Jaccard from
0.101 to 0.294, and fine-label F1 from 0.141 to 0.418 while every binary verdict
remains fixed at 0.930 accuracy. The paired Jaccard gain is $+0.193$ (95\%
bootstrap CI $[0.134,0.254]$), demonstrating that the improvement comes from
recovering supported failure causes rather than shifting the decision
threshold.

\noindent\textbf{The resulting advice transfers and supports efficient expert
triage.}
On the independent cross-suite panel, the final audit supplies at least one
expert-supported cause for 84\% of erroneous structures, versus 60\% for the
sealed first pass; Jaccard also rises by $+0.091$ (95\% CI $[0.032,0.155]$).
Chem-GD identifies expert-correct causes with 0.938 precision and 0.714 recall.
Operationally, each accepted audit names a likely failure mode and points to
specific structural evidence, turning open-ended inspection of a long CIF into
targeted verification of a bounded recommendation. This supports practical
curation efficiency without outsourcing the final chemical judgment.

\begin{table}[H]
\centering
\scriptsize
\makebox[\columnwidth][l]{\textbf{(a) Blind verdict reliability}}\par\vspace{3pt}
\setlength{\tabcolsep}{4.2pt}
\renewcommand{\arraystretch}{1.05}
\begin{tabular}{@{}>{\raggedright\arraybackslash}p{0.50\columnwidth}
                    >{\centering\arraybackslash}p{0.205\columnwidth}
                    >{\centering\arraybackslash}p{0.205\columnwidth}@{}}
\toprule
Comparison & Agreement & Cohen's $\kappa$ \\
\midrule
\rowcolor{groupbg}
Expert A vs.\ Expert B & 0.980 & 0.960 \\
\rowcolor{groupbg}
Database label vs.\ consensus & 1.000 & 1.000 \\
\rowcolor{oursbg}
\textbf{\textsc{MOF-Sleuth} vs.\ consensus} & \textbf{0.930} & \textbf{0.860} \\
\bottomrule
\end{tabular}

\vspace{7pt}
\makebox[\columnwidth][l]{\textbf{(b) Fine-grained attribution on erroneous cases}}\par\vspace{3pt}
\setlength{\tabcolsep}{4.2pt}
\renewcommand{\arraystretch}{1.05}
\begin{tabular}{@{}>{\raggedright\arraybackslash}p{0.43\columnwidth}
                    >{\centering\arraybackslash}p{0.15\columnwidth}
                    >{\centering\arraybackslash}p{0.15\columnwidth}
                    >{\centering\arraybackslash}p{0.15\columnwidth}@{}}
\toprule
Comparison & Type-Hit Acc & Jaccard & Fine-label F1 \\
\midrule
\rowcolor{groupbg}
Expert A vs.\ Expert B & 0.480 & 0.252 & 0.389 \\
\rowcolor{groupbg}
Expert A vs.\ consensus & 0.960 & 0.528 & 0.635 \\
\rowcolor{groupbg}
Expert B vs.\ consensus & 1.000 & 0.702 & 0.839 \\
\rowcolor{oursbg}
\textbf{\textsc{MOF-Sleuth} vs.\ consensus} & \textbf{0.700} & \textbf{0.294} & \textbf{0.418} \\
\bottomrule
\end{tabular}

\vspace{7pt}
\makebox[\columnwidth][l]{\textbf{(c) Conditional refinement and cross-suite transfer}}\par\vspace{3pt}
\setlength{\tabcolsep}{3.8pt}
\renewcommand{\arraystretch}{1.08}
\begin{tabular}{@{}>{\raggedright\arraybackslash}p{0.25\columnwidth}
                    >{\centering\arraybackslash}p{0.335\columnwidth}
                    >{\centering\arraybackslash}p{0.335\columnwidth}@{}}
\toprule
Metric & ToBaCCo-100 & Cross-suite-100 \\
\midrule
\rowcolor{oursbg}
Type-Hit Acc &
$0.260\!\to\!\mathbf{0.700}$ \textcolor{okgreen}{(+0.440)} &
$0.600\!\to\!\mathbf{0.840}$ \textcolor{okgreen}{(+0.240)} \\
\rowcolor{oursbg}
Error-case Jaccard &
$0.101\!\to\!\mathbf{0.294}$ \textcolor{okgreen}{(+0.193)} &
$0.396\!\to\!\mathbf{0.487}$ \textcolor{okgreen}{(+0.091)} \\
\rowcolor{oursbg}
Fine-label F1 &
$0.141\!\to\!\mathbf{0.418}$ \textcolor{okgreen}{(+0.277)} &
$0.219\!\to\!\mathbf{0.324}$ \textcolor{okgreen}{(+0.105)} \\
\rowcolor{oursbg}
Chem-GD agreement &
$0.700\!\to\!\mathbf{0.800}$ \textcolor{okgreen}{(+0.100)} &
$0.600\!\to\!\mathbf{0.720}$ \textcolor{okgreen}{(+0.120)} \\
\bottomrule
\end{tabular}
\caption{Blinded expert validation of verdict reliability, fine-grained
attribution, and cross-suite transfer. Panel (a) uses all ToBaCCo-100
cases, and panel (b) uses its adjudicated erroneous cases. In (c), ``single''
is the sealed first-pass output and ``final'' is conditional attribution after
freezing the protocol; verdicts never change. Type-Hit Acc requires at least
one predicted fine type to match an expert-supported cause, and Chem-GD
agreement compares automatic pass/fail with expert attribution correctness.}
\label{tab:human-validation}
\label{tab:human-eval}
\label{tab:human-attribution-comparison}
\label{tab:expert-attribution}
\end{table}

\appendixanchor{appendix-demo}
\section{Interactive Audit Demo}
\label{app:demo}

We ship the framework as a local HTML demo. A dropped CIF is parsed locally,
an evidence report is compiled from the Forensic Lab checks, and only that
report is sent to a user-configured LLM endpoint under the paper's
evidence-report prompt. The returned audit is parsed into the
reasoning\,/\,\texttt{error\_type}\,/\,\texttt{answer} schema, and each cited
field--value pair is checked against the locally computed tool values. The
following three pages show one complete walkthrough on the built-in sample
case: case intake and tool-by-tool forensics, the compiled evidence report, the
interrogation configuration, and the structured verdict. The examiner output
shown is the verbatim GPT-5.5 response (API key masked, gateway URL replaced by
a placeholder).

\onecolumn
\begin{center}
\includegraphics[page=1,width=0.71\textwidth]{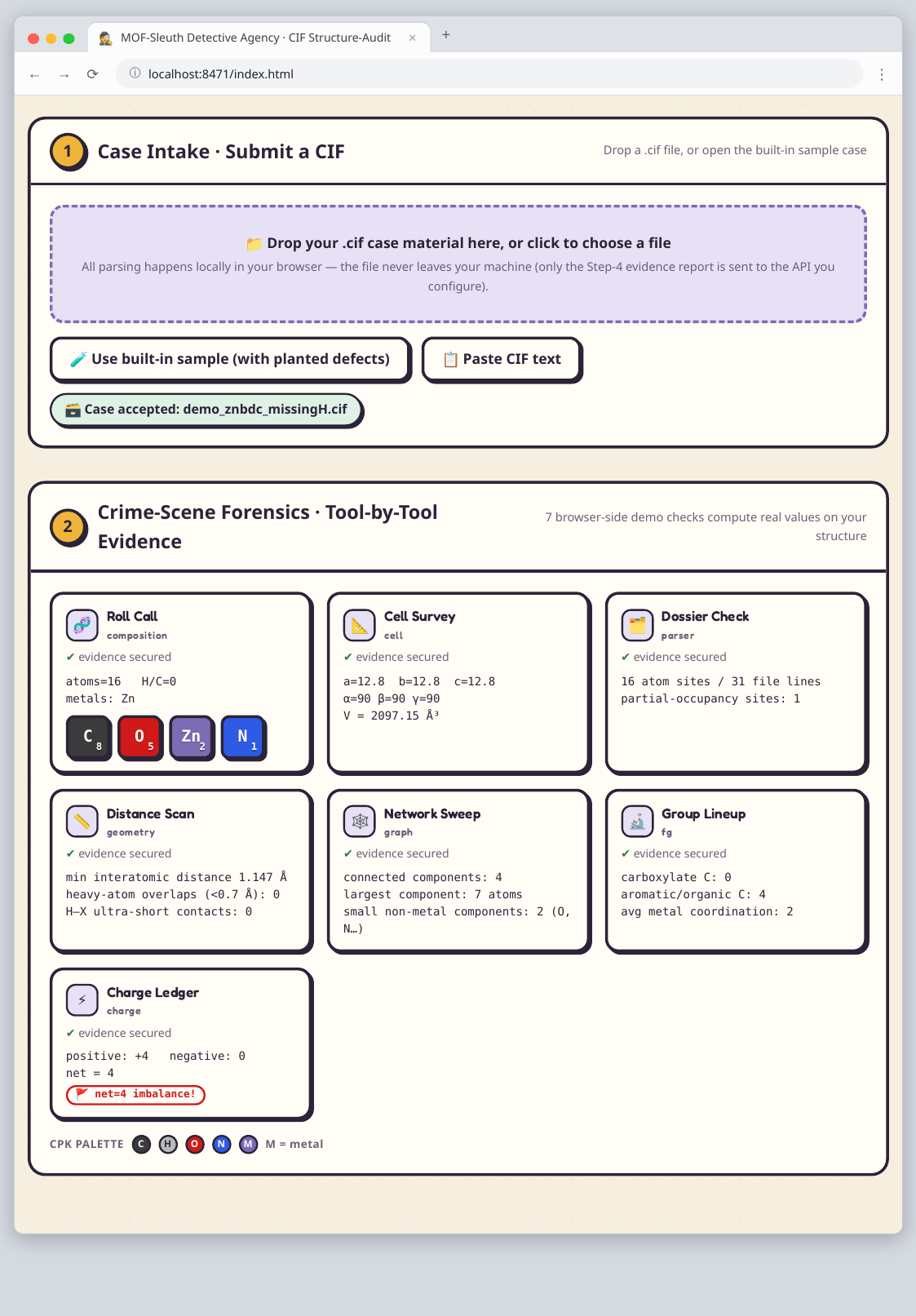}
\end{center}
\begin{center}
\includegraphics[page=2,width=0.71\textwidth]{Figures/demo_walkthrough.pdf}
\end{center}
\begin{center}
\includegraphics[page=3,width=0.71\textwidth]{Figures/demo_walkthrough.pdf}
\end{center}

\end{document}